\documentclass[lettersize,journal]{IEEEtran}
\usepackage{amsmath,amsfonts}
\usepackage{algorithmic}
\usepackage{algorithm}
\usepackage{array}
\usepackage{textcomp}
\usepackage{stfloats}
\usepackage{url}
\usepackage{verbatim}
\usepackage{graphicx}
\usepackage{cite}
\usepackage{adjustbox}
\usepackage{booktabs}
\usepackage{caption}
\usepackage{colortbl}
\usepackage{rotating}
\usepackage{multirow}
\usepackage{soul}
\usepackage{tabularx}
\usepackage{threeparttable}
\usepackage{xcolor}
\usepackage{xtab}
\usepackage{subfigure} 
\usepackage{authblk}
\usepackage{makecell}
\usepackage{hyperref}
\hypersetup{hidelinks,
	colorlinks=true,
	allcolors=black,
	pdfstartview=Fit,
	breaklinks=true}

\begin{document}

\newcommand{\hy}[1]{\textcolor{red}{#1}}

\title{FuseGrasp: Radar-Camera Fusion for Robotic
Grasping of Transparent Objects}

\author{Hongyu Deng, Tianfan Xue, and He (Henry) Chen,~\IEEEmembership{Member,~IEEE}
\thanks{Hongyu Deng, Tianfan Xue and He (Henry) Chen are with the Department of Information Engineering, The Chinese University of Hong Kong, Hong Kong SAR, China. {E-mail: \{dh021, tfxue, he.chen\}@ie.cuhk.edu.hk}.}

\thanks{He (Henry) Chen is also with the T Stone Robotics Institute, The Chinese University of Hong Kong, Hong Kong SAR, China.}
}

\maketitle

\begin{abstract}
Transparent objects are prevalent in everyday environments, but their distinct physical properties pose significant challenges for camera-guided robotic arms. Current research is mainly dependent on camera-only approaches, which often falter in suboptimal conditions, such as low-light environments. In response to this challenge, we present FuseGrasp, the first radar-camera fusion system tailored to enhance the transparent objects manipulation. FuseGrasp exploits the weak penetrating property of millimeter-wave (mmWave) signals, which causes transparent materials to appear opaque, and combines it with the precise motion control of a robotic arm to acquire high-quality mmWave radar images of transparent objects. The system employs a carefully designed deep neural network to fuse radar and camera imagery, thereby improving depth completion and elevating the success rate of object grasping. Nevertheless, training FuseGrasp effectively is non-trivial, due to limited radar image datasets for transparent objects. We address this issue utilizing large RGB-D dataset, and propose an effective two-stage training approach: we first pre-train FuseGrasp on a large public RGB-D dataset of transparent objects, then fine-tune it on a self-built small RGB-D-Radar dataset. Furthermore, as a byproduct, FuseGrasp can determine the composition of transparent objects, such as glass or plastic, leveraging the material identification capability of mmWave radar. This identification result facilitates the robotic arm in modulating its grip force appropriately. Extensive testing reveals that FuseGrasp significantly improves the accuracy of depth reconstruction and material identification for transparent objects. Moreover, real-world robotic trials have confirmed that FuseGrasp markedly enhances the handling of transparent items. A video demonstration of FuseGrasp is available at  \url{https://youtu.be/MWDqv0sRSok}.
\end{abstract}

\begin{IEEEkeywords}
Transparent object grasping, robotic arm, radar-camera fusion, synthetic aperture radar (SAR), material identification.
\end{IEEEkeywords}

\vspace{-1em}
\section{Introduction}\label{sec_introduction}

\begin{figure}[]
	\centering	
	\includegraphics[width = 0.85\linewidth]{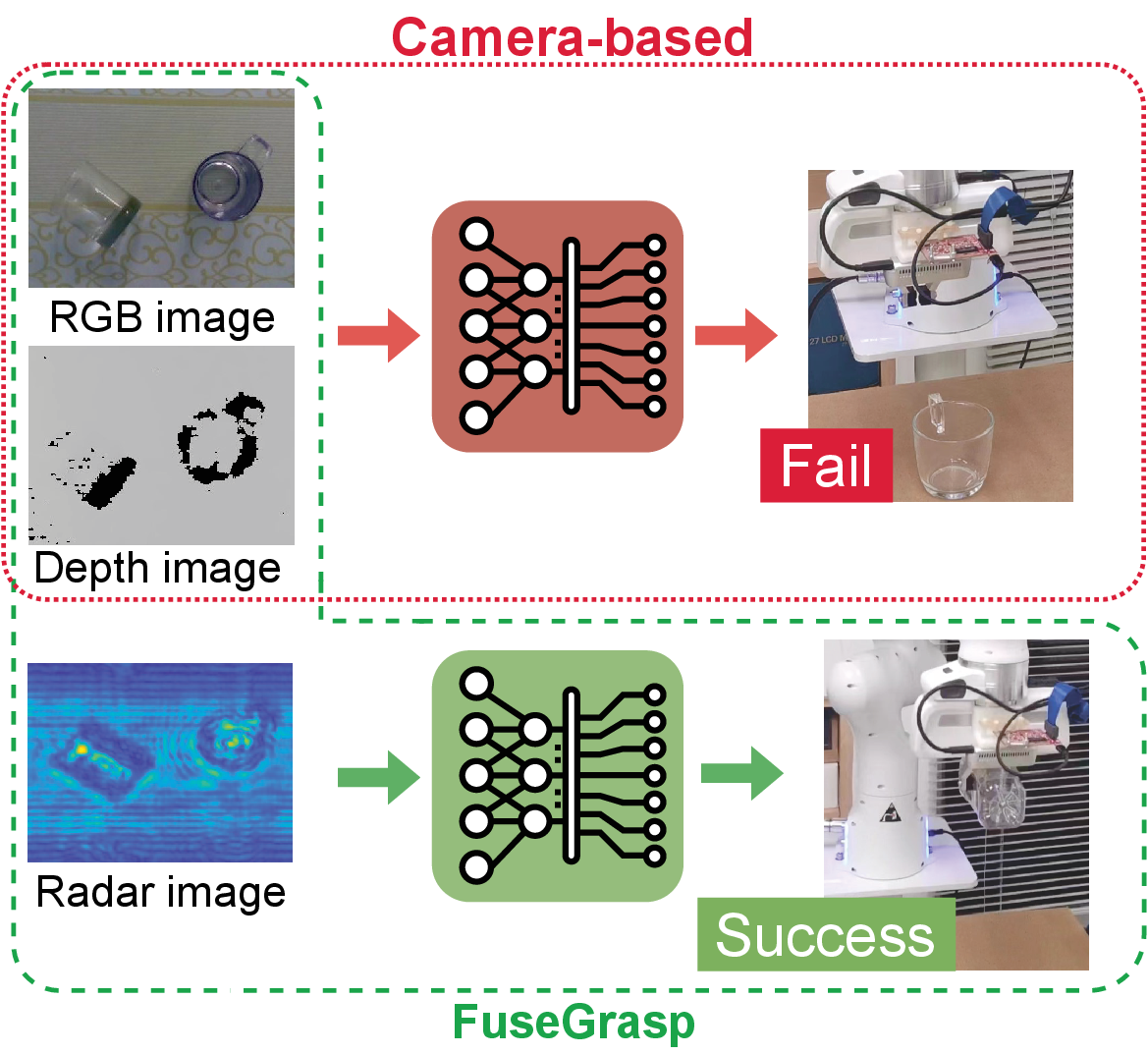} 
	\caption{
		\textbf{\emph{Top.}} Camera-based solutions using RGB-D imagery often face challenges in accurately detecting and grasping transparent objects under variable lighting conditions.
		\textbf{\emph{Bottom.}} FuseGrasp strategically fuses RGB-D and radar imagery to enhance the precision and reliability of robotic grasping when handling transparent objects.
	}	
	\label{motivation_rgb_depth}
	\vspace{-1em}
\end{figure}

Robotic arms, a critical segment of robotics, are extensively utilized in sectors like manufacturing and domestic services. Although modern robotic arms can effectively locate and handle items in various work environments using computer vision, they falter when encountering transparent materials. RGB-D cameras, which depend on the reflection of light paths, are adept at recognizing optically opaque objects but struggle with transparent ones such as glass and plastic—commonplace yet complex in everyday life \cite{transcg}. The refractive and reflective properties of transparent objects challenge conventional geometric light path assumptions, thereby complicating the task of accurate depth estimation. Furthermore, the appearance of transparent materials can drastically change depending on the background, which poses difficulties for standard visual detection methods. 
For instance, the depth perception errors evident in the depth image shown in Fig. \ref{motivation_rgb_depth}, taken under normal lighting conditions, highlight this issue. 
Our findings align with those reported in the broader research community, as evidenced by studies such as \cite{cleargrasp, keypose, zhu2021rgb}.

Despite recent advances by researchers, the inherent properties of transparent objects continue to pose a formidable challenge in their detection and manipulation within the field of robotics, as highlighted in a recent survey \cite{jiang2023robotic}. Existing literature documents various vision-based techniques developed to pinpoint the location, shape, and material composition of transparent objects \cite{albrecht2013seeing, phillips2016seeing, zhou2020lit, zhu2015reusing}. These methods employ neural networks to identify features and correct depth perception errors, thus facilitating the reconstruction of transparent objects' geometries \cite{transcg, stereopose, dex-nerf, evo-nerf, depthgrasp}. Yet, these vision-centric solutions often underperform in adverse conditions such as low light or cluttered backgrounds. To compensate, some researchers have integrated supplementary sensors with traditional cameras, capitalizing on the rich data provided by RGB-D technology \cite{kalra2020deep, RF-grasp, RFusion}. For example, PokePreNet is the first work to fuse vision and tactile sensors for transparent object detection, which trains neural networks using synthetic datasets to enhance localization accuracy \cite{jiang2022shall}. Another visual-tactile approach has been suggested for identifying potential glass objects, even in aquatic environments \cite{visual_tactile}. However, these alternative sensors can be cumbersome in searching across operational planes, may not seamlessly mesh with visual data, and often come with a high cost. Addressing these limitations necessitates the adoption of a more suitable sensor to augment RGB-D capabilities for more effective localization and gripping of transparent objects.

Radar sensors represent a promising class of wireless devices for object detection, capable of providing accurate distance measurements even in challenging environments \cite{kouros20183d, zheng2021more, zheng2021siwa, zhang2023rf, ref1_yao2023radar, ref3_yao2023radar}. Within the radar sensor family, mmWave radar, equipped with multiple transceivers, has garnered significant interest \cite{wei2015mtrack, lu2020see, lu2020milliego}. Distinct from lower frequency signals, mmWave signals possess shorter wavelengths that are more susceptible to scattering. This characteristic, though a hindrance for mmWave communication, can be beneficial for detecting transparent objects; the shorter wavelengths of mmWave signals are less penetrating, making transparent materials appear opaque at these frequencies. Moreover, the increased scattering from smooth surfaces enables radar to receive more reflected signals in indoor environment \cite{smooth_reflect1, smooth_reflect2}. Consequently, fusing camera imagery with mmWave radar data is a convincing solution for the robotic grasping of transparent objects.

However, developing a camera-radar fusion system for a robotic arm tasked with grasping transparent objects also introduces several challenges. Firstly, radar sensors capture sparse echoes that are often contaminated by multipath interference. While initial signal processing efforts can suppress noise, they are insufficient for reconstructing the shape of the target object. An advanced technique for generating accurate radar images of transparent items is therefore essential. Secondly, the process of collecting radar data for this purpose is labor-intensive, and there is a dearth of open-source radar datasets dedicated to transparent objects. The scarcity of data hampers the ability to train a comprehensive end-to-end neural network model for transparent object reconstruction. Lastly, the disparate nature of visual and radar imagery, each with its own resolution characteristics, calls for a method to extract and correlate features from both modalities effectively. It is therefore imperative to conceive a fusion model that can seamlessly combine the information from camera and radar sources to support the robotic grasping function.

To overcome these hurdles, we introduce FuseGrasp, the first robotic system designed to \underline{fuse} radar-camera data for enhanced transparent object \underline{grasp}ing, as illustrated in Fig. \ref{motivation_rgb_depth}. FuseGrasp skillfully harnesses the precise movement capabilities of the robotic arm to create an expansive synthetic antenna array. Utilizing synthetic aperture radar (SAR) technology, it processes the received signals to generate accurate radar images. These images are used to complement RGB-D visuals for enhanced depth completion, thereby enhancing the detection and grasping accuracy of transparent objects. In response to the scarcity of publicly available radar image datasets for transparent objects and the laborious task of constructing a large dataset independently, FuseGrasp strategically employs a two-stage deep learning training approach. Initially, we utilize a large open-source RGB-D dataset to preliminarily extract features of transparent objects. Subsequently, we refine the model using a small, self-built dataset that integrates RGB-D with corresponding radar data, leveraging the complementary strengths of these modalities to boost detection and grasping performance. Additionally, FuseGrasp offers the supplementary advantage of modulating the grasping force through the material recognition ability of the mmWave radar, allowing for nuanced adjustments such as applying less force for soft plastic compared to glass. 
Real-world robotic arm experiments confirm that FuseGrasp markedly improves the grasping of transparent objects, proving its efficacy across a spectrum of lighting scenarios.

\vspace{-0.5em}
\section{Background and Feasibility Study}\label{sec_model}
This section begins with an introduction to the basics of mmWave radar signals and the principles of SAR imaging using mmWave radar. We then demonstrate the feasibility of fusing mmWave radar and camera data while highlighting the challenges associated with this approach.

\subsection{mmWave Radar Basics and SAR Imaging}

\subsubsection{mmWave Radar Basics}
\begin{figure}[]
	\centering
	\includegraphics[width=0.95\linewidth]{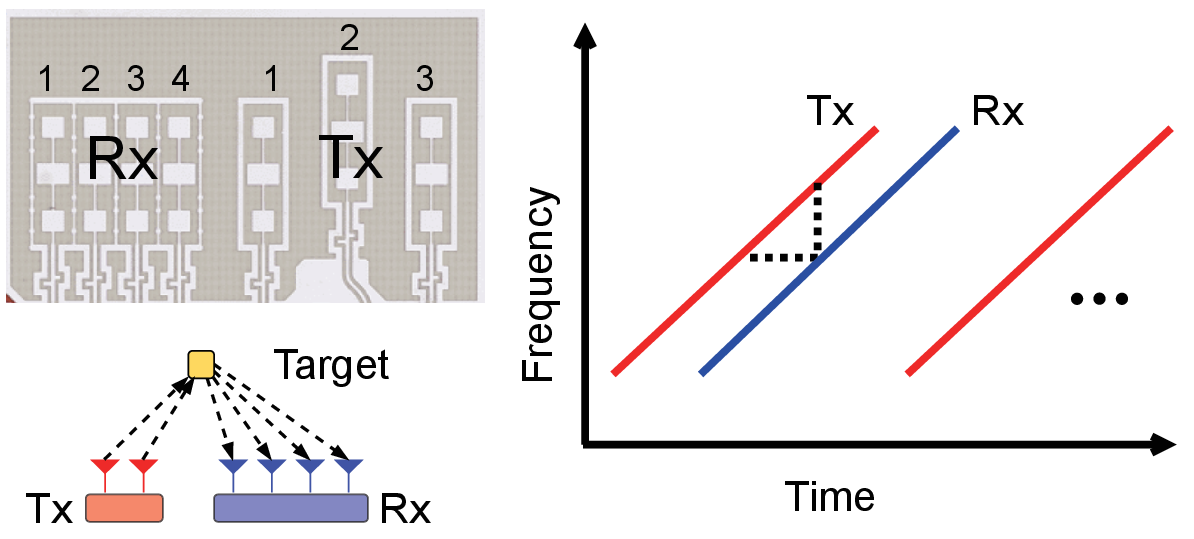}
	\caption{
		\textbf{\emph{Left.}} Physical antenna array layouts on mmWave radar boards, and the $2 \times 4$ Tx-Rx pairs used in FuseGrasp. 
		\textbf{\emph{Right.}} Sequence of transmitted signals (red) and received signals (blue).}
	\vspace{-1em}
	\label{mmWave}
\end{figure}

The mmWave radar employs multiple-input multiple-output (MIMO) technology to receive multi-channel echoes simultaneously. We illustrate this process in Fig. \ref{mmWave}. The left figure depicts the physical antenna layouts used on the adopted mmWave radar. The radar generates and transmits a series of linearly frequency-modulated waves, and the received echoes can be used to estimate the range of reflectors. In an ideal scenario, the unit-power signal emitted can be represented in complex notation~as
\begin{equation}\label{eqn1}
	\begin{aligned}
		s(t) = e^{j 2 \pi \left(f_0 t+ 0.5 K t^2\right)},
	\end{aligned}
\end{equation}
where $t$ represents time within the interval $[0, T]$, with $T$ denoting the duration of a single chirp wave. $K = \frac{B}{T}$ is the frequency slope, where $B$ represents the bandwidth. $f_0$ and $f_0 + B$ signify the starting and ending frequencies, respectively.

The received (Rx) signal is a time-delayed version of the transmitted (Tx) signal. The mmWave radar can measure the corresponding time delay $\tau$ from the intermediate frequency signal. The complex intermediate frequency signal can be expressed as
\begin{equation}\label{eqn2}
	\begin{aligned}
		r(t) = g_r e^{j 2 \pi (f_0 \tau + K t \tau - 0.5K \tau^2)} \approx g_r e^{j 2 \pi {(f_0  + K t)\tau}},
	\end{aligned}
\end{equation}
where $g_r$ {denotes the complex scattering coefficient}, and $\tau K = \frac{2d}{c}K$ represents the frequency corresponding to the object distance $d$, with $c$ being the speed of light. The second-order term, known as the residual video phase, has been shown to be negligible in prior research \cite{RVP}.

\subsubsection{SAR Imaging}
{The complex coefficient} $g_r$ in Eq. (\ref{eqn2}) denotes the backscattering coefficient of a point target, which is dependent on the surface roughness and the dielectric constant of the reflector. Typically, a pair of transceivers can collect only a narrow range of reflected signals. To obtain echo data over a wider area, the deployment of a large antenna array is usually necessary. To accomplish this, we exploit the precise movement of the robotic arm to emulate a large antenna array and generate SAR images. SAR imaging, unlike other spatial signal processing methods, not only determines the precise coordinates but also yields the geometric information of the detected object.

\begin{figure}[]
	\centering
	\includegraphics[width=0.95\linewidth]{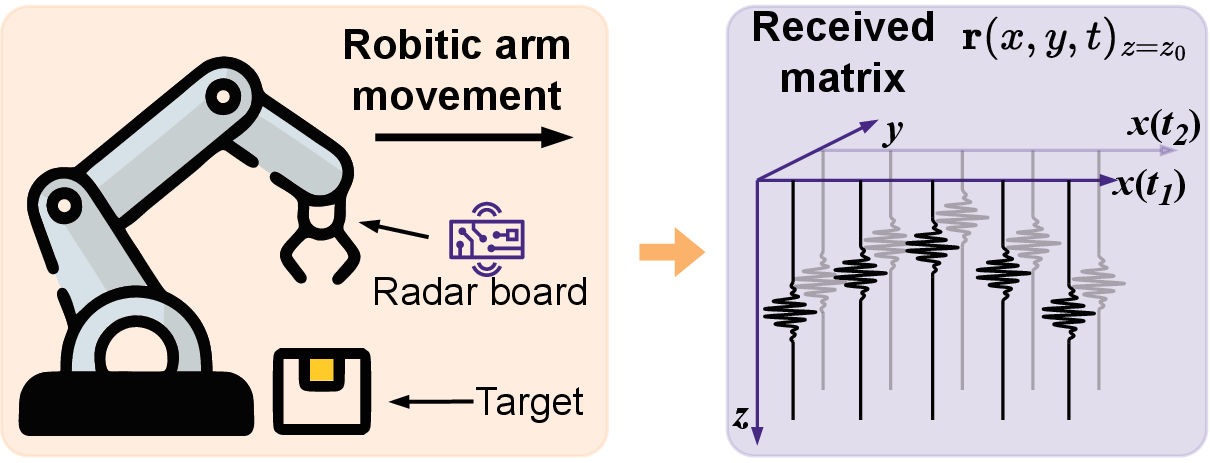}
	\caption{\textbf{Formation of received signal.} With the movement of the robotic arm, the mmWave radar attached to the hand will generate a virtual antenna array. By jointly processing the received signal at each antenna element, radar images of the detected targets can be synthesized.}
	\label{formation}
\end{figure}

In particular, we adopt the range migration algorithm (RMA) to synthesize high-quality radar images. RMA stands out among SAR imaging methodologies for its ability to effectively balance imaging resolution with computational efficiency \cite{UT1}. Given that our robotic arm maneuvers the mmWave sensor along both the $x$ and $y$ axes, as depicted in Fig. \ref{formation}, we have adapted the SAR imaging algorithm for two-dimensional application. In the context of RMA \cite{3DRIED}, the received intermediate frequency signal, as defined in Eq. (\ref{eqn2}), is transformed into the wave number domain given by
\begin{equation}\label{eqn3}
	\begin{aligned}
		r(k) = g_r e^{j2kd}, \quad \tfrac{2 \pi f_0}{c} \leq k \leq \tfrac{2 \pi f_0+KT}{c},
	\end{aligned}
\end{equation}
where $k$ is the wave number, representing the phase change in the spatial domain.

{As delineated by the radar equation, signal amplitude is subject to variation with distance \cite{he2023fusang}. Nonetheless, due to the near-field operating conditions of our system, we will disregard the propagation effects attributable to distance. The position of our operational platform is fixed, allowing for the prior determination of the $z$-axis coordinate in the target's location, represented by $z_0$. Consequently, Eq. (\ref{eqn3}) can be extended to 2-D as follows:}
\begin{equation}\label{eqn4}
	\begin{aligned}
		r(x,y,k) = \iint g_r(x,y,z_0) e^{j2kd(x,y,z_0)} \, dx\,dy ,
	\end{aligned}
\end{equation}
where $d(x,y,z_0)$ is the distance between the transceiver and the target point at coordinate $(x,y,z_0)$.
RMA uses the 2-D Fourier transform (FT) and inverse Fourier transform (IFT) to obtain the complex target reflectivity $g_r(x,y)$ at the plane $z = z_0$  \cite{3DRIED}. Mathematically, we have 

\begin{equation}\label{eqn8}
	\begin{aligned}
		g_r(x,y) = \int \text{IFT}_{2D} \Big[ \text{FT}_{2D} \big[r(x,y,k) \big]e^{-j k_z z_0} \Big] \, dk,
	\end{aligned}
\end{equation}
where $k_z = \sqrt{4k^2 - k^2_x - k^2_y}$, and $k_x$, $k_y$, $k_z$ are the components of the spatial wave number $k$ in $x$, $y$, and $z$ dimensions, respectively. Finally, the pixel value of the RMA image at the coordinate $(x,y)$ can be obtained by computing the magnitude of $g_r(x,y)$, i.e., $\left|g_r(x,y)\right|$.

\vspace{-0.5em}
\subsection{Feasibility and Challenges of Radar-Camera Fusion}

\begin{figure}[]
	\centering	
	\includegraphics[width = 1.02\linewidth]{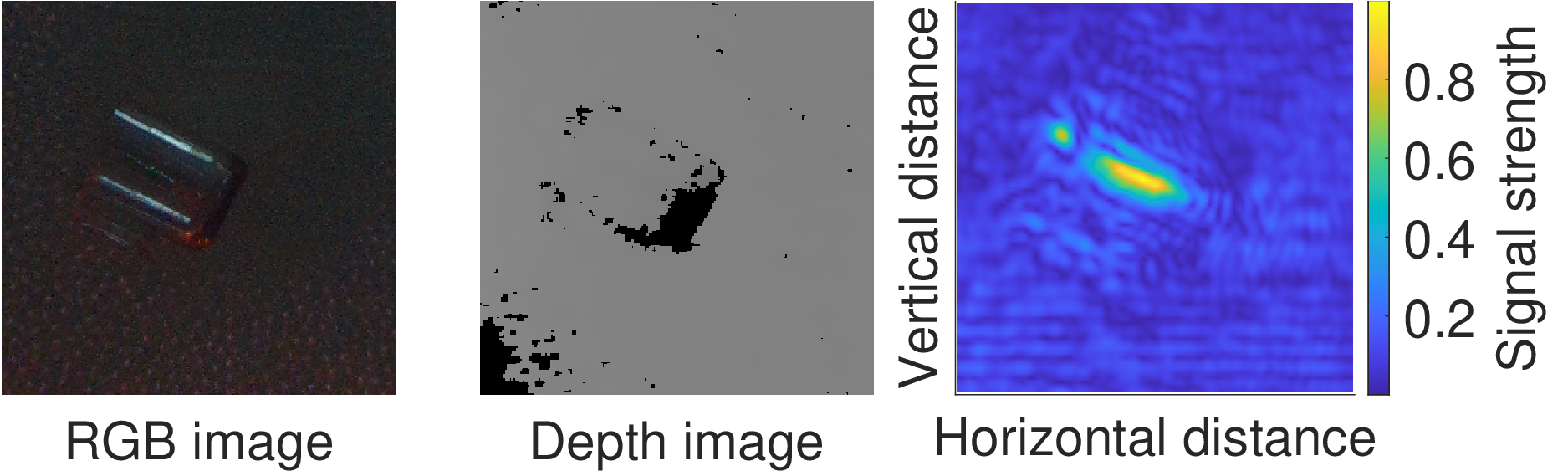} 
	\caption{Visualization of the preliminary RMA radar imaging (right) of one cup under dark conditions and the corresponding RGB image (left) and depth image (middle).}
	\label{motivation_SAR}
	\vspace{-1em}
\end{figure}

To demonstrate the feasibility of integrating radar imaging with RGB-D imagery, we have produced a SAR image of one cup in dark conditions, which is presented in Fig. \ref{motivation_SAR} together with its RGB and depth counterparts. The SAR image on the right in Fig. \ref{motivation_SAR} confirms that radar imagery can effectively outline transparent objects, irrespective of lighting conditions. Such images can act as auxiliary information to augment RGB-D imagery, thereby substantiating the potential of integrating radar and camera data to accurately characterize transparent objects.

Nevertheless, accurately reconstructing the shape of transparent objects by fusing radar and RGB-D imagery remains a considerable challenge, due to the limited training data. Specifically, while RGB-D data for transparent objects is widely available, there is no public radar dataset for such objects. Collecting a large RGB-D-Radar dataset for transparent objects is particularly difficult due to the intensive labor in capturing radar data. In light of this, we ask a fundamental question: Can we harness mmWave radar's capability in characterizing transparent objects by leveraging the existing RGB-D datasets alongside a small RGB-D-Radar dataset? The asymmetry of RGB-D and radar datasets makes direct fusion models ineffective. As shown in Table~\ref{total_result} of Sec.~\ref{sec_evaluation}, a direct fusion of RGB-D and radar data leads to marginal gains or even performance degradation. This necessitates the development of a new deep learning approach that can effectively merge radar and RGB-D data.
In the next section, we will outline the design elements of FuseGrasp and explain how it addresses the aforementioned issue.

\begin{figure*}[]
	\centering
	\includegraphics[width= 0.95\textwidth]{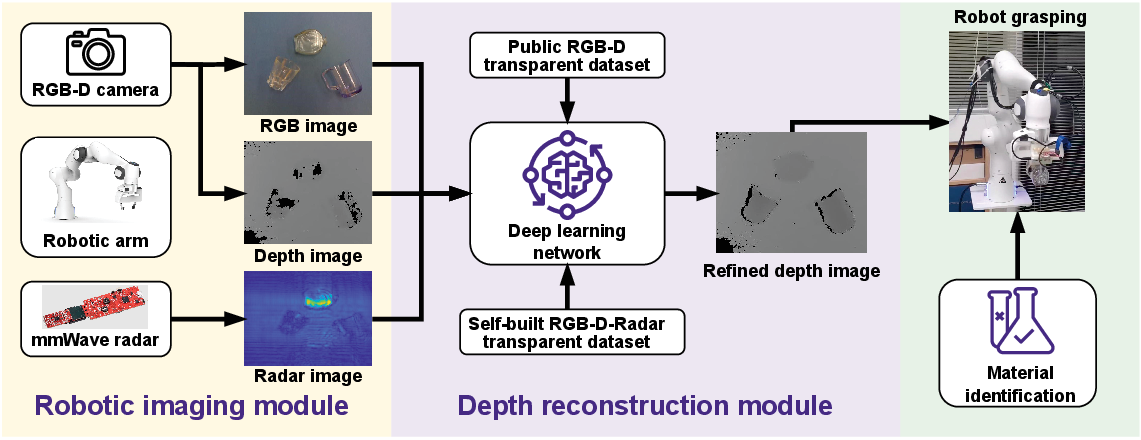}
	\caption{\textbf{The architecture of FuseGrasp.} It integrates an RGB-D camera and mmWave radar through four key modules: robotic imaging, depth reconstruction, material identification, and grasping.}
	\vspace{-1em}
	\label{Overview}
\end{figure*}

\section{FuseGrasp Design} \label{sec_sys_design}
This section presents the architecture of FuseGrasp, as depicted in Fig. \ref{Overview}. The system is composed of four interconnected modules: robotic imaging, depth reconstruction, material identification, and grasping. The robotic imaging module utilizes SAR imaging to produce high-quality radar images of transparent objects. For depth reconstruction, a neural network—initially trained on publicly available RGB-D datasets and subsequently fine-tuned using our own RGB-D-Radar dataset—deduces the shape of objects from camera and radar images. In the material identification module, the type of material an object is made from is determined using radar data. Finally, the grasping module orchestrates the movements of the robotic arm, enabling it to autonomously grasp objects as informed by the data processed in the preceding modules.

\subsection{Robotic Imaging Module}\label{sec:imgage}
\begin{figure}[]
	\centering
	\includegraphics[width= 0.47\textwidth]{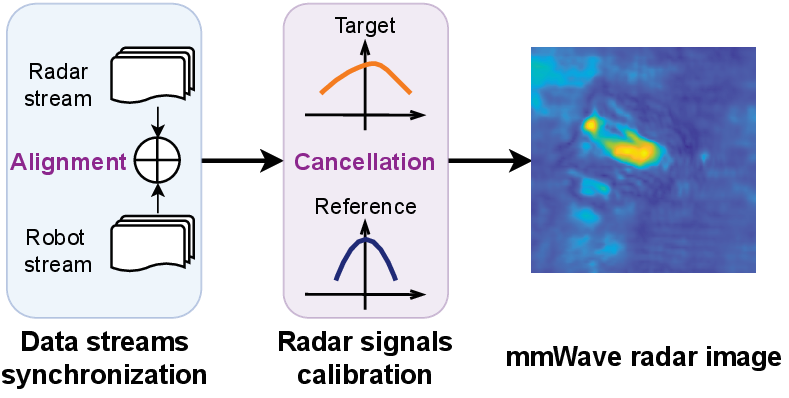}
	\caption{\textbf{Robotic radar imaging pipeline.} The radar image is an improved version to the one shown in Fig.~\ref{motivation_SAR}.}
	\vspace{-1em}
	\label{Pipeline}
\end{figure}

This module performs both camera and radar imaging. {The  RGB-D images are captured at the initial position of the robot hand.} For the radar imaging, the radar mounted on the robotic arm is employed to capture SAR images of transparent objects. The radar imaging module is structured around two main operations: \textit{a) Data Streams Synchronization}: We utilize a unified clock reference to bring two distinct data streams—the received mmWave signal and the robotic arm's position signal—into temporal alignment. By synchronizing these data streams, we significantly reduce distortions in the radar imagery. b) \textit{Radar Signals Calibration}: With the data streams accurately synchronized, we proceed to calibrate any discrepancies in the radar signals. This calibration process is crucial for improving the fidelity of the radar images. The workflow of the robotic radar imaging module, including these two steps, is depicted in Fig.~\ref{Pipeline}.

\subsubsection{Data Streams Synchronization} 
To capture high-quality radar images, a primary challenge is the synchronization of the time-clocks between the mmWave radar signal and the robotic arm's position signal. {In FuseGrasp, the received radar signal and the robotic arm positional signal are concurrently collected.} Both systems are wired to a central workstation PC, enabling us to treat the configuration as a master-slave setup. For synchronization purposes, we employ the C++ Chrono library. The \verb|system_clock| class in this library taps into the system-wide real-time clock, which is capable of directly accessing the hardware clock \cite{Chrono}. Our methodology involves having both devices issue a timestamp with each data transmission to the workstation. Empirical analyses reveal that the robotic arm system has an average data round-trip latency of 0.24 ms, while the mmWave device exhibits an average time delay of 0.16 ms. Given the radar's low frame rate and the robotic arm's smooth movement, these minor discrepancies are negligible in practice. This ensures that time records are dependable and synchronization is sufficiently precise for our application.

Before further process, we first need to temporally align two data streams. This is because there is a gap between the sampling rates of the two devices, even though their timestamps are recorded. In particular, the robotic arm provides position data at a default rate of 1000 frames per second (FPS), which is much higher than the mmWave radar's rate of 80 FPS. Additionally, the mmWave radar starts collecting data continuously from the moment it is turned on, which is usually earlier than the robotic arm starts. This results in the two sets of data having different starting points. Because of these differences, the timestamps from the two devices do not line up, and we have to remove any data that does not correspond in time between the two streams.
To address this issue, we use the initial and final timestamps of the robotic arm as the start and end times, treating the less-sampled mmWave radar sample points as the reference time sequence. 
During the motion of the arm, given any location coordinate of the virtual antenna array as well as the corresponding timestamp in the robotic arm position data stream, we determine the nearest timestamp in the corresponding radar data stream. By retrieving the corresponding received radar signal at such timestamp, we can construct the received radar signal matrix, as illustrated in Fig.~\ref{formation}.

\begin{figure}[t]
	\centering	
	\subfigure[Radar images of point target.]{\includegraphics[width = 0.505\linewidth]{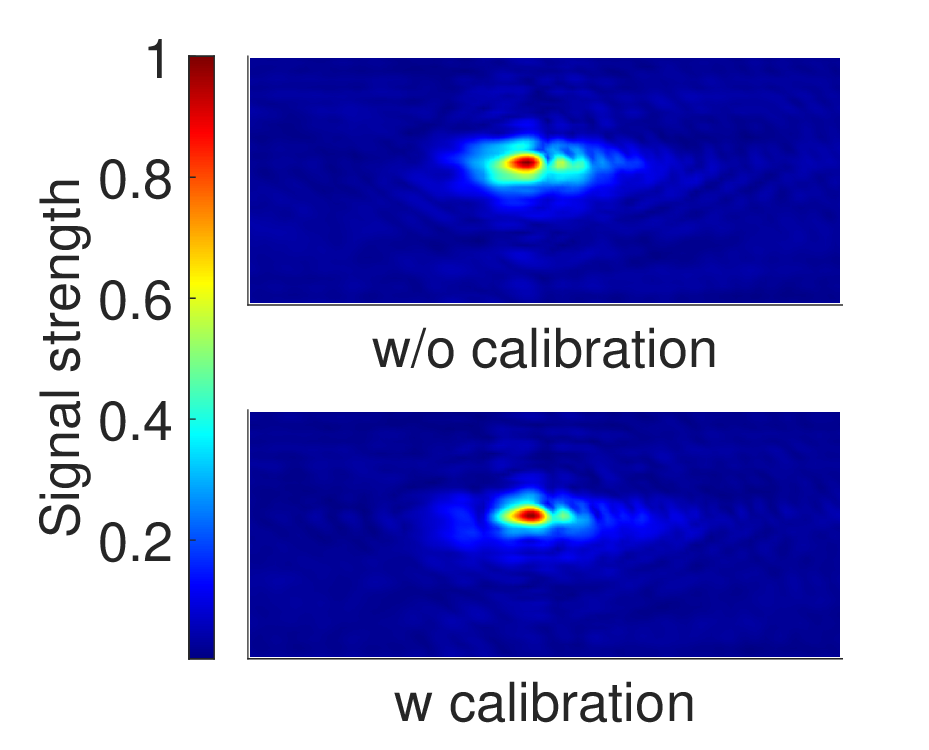}}\label{Calibration_target_radar}				
	\subfigure[Amplitude profiles of point target.]{\includegraphics[width = 0.465\linewidth]{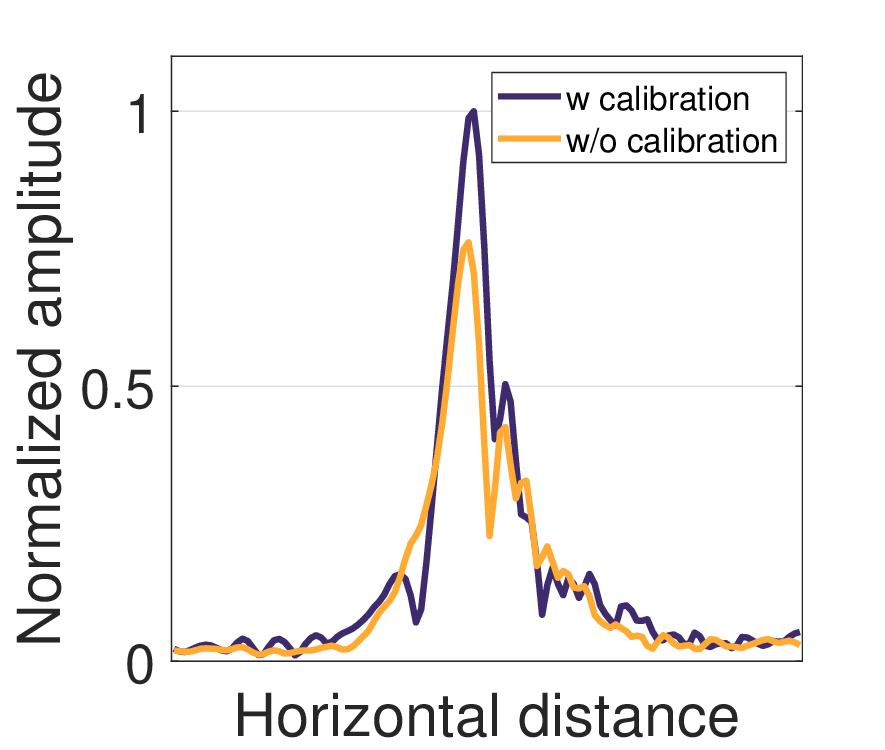}}
	\label{Calibration_target_amp}
	
	\caption{\textbf{Radar signal calibration for a point target.} In an ideal case, the amplitude of the signal profiles as a function of horizontal distance is expected to exhibit the shape of a sinc function.}\label{Calibration}
	\vspace{-1em}
\end{figure}

\subsubsection{{Radar Signal Calibration}}
In real-world scenarios, factors such as channel mismatches, hardware offsets, and background noise can lead to phase errors, which in turn may result in image blurring and the presence of ghost artifacts, ultimately compromising the quality of radar imaging. Consequently, the second critical step in improving image quality entails the calibration of signal offsets to mitigate these issues.

Previous studies have outlined a range of calibration techniques \cite{Calib1, Calib2}. Yet, these methods often prove to be complex and challenging to implement in operational settings. We introduce a streamlined calibration method designed to overcome these practical hurdles. Our approach hinges on using a static point target with high reflectivity as a benchmark to neutralize signal offsets. To accomplish this, we must first develop a model of the reference signal that accurately represents the object's precise location. The initial step in our calibration process involves determining the unknown coordinates $(\hat{x}, \hat{y}, \hat{z})$ of the reflective object's point of reflection. 

Assuming there are $K$ sets of data captured at different locations, we can leverage position signal from the robotic arm to ascertain the precise coordinates $(x_k, y_k, z_k)$ of the radar's $k$-th position. It becomes clear that the problem of signal calibration can be reframed as a problem of triangulating the three-dimensional position of the reflection point. Given that the mmWave radar can measure the distance $d_k$ between the reflective object and the $k$-th observation point directly, we can establish the relationship $(x_k - \hat{x})^2 + (y_k - \hat{y})^2 + (z_k - \hat{z})^2 = d_k^2$. Although there may be some offsets in the measured distance, these can be mitigated by averaging the distances reported by adjacent antennas. With this averaged measurement, we can then determine the unknown coordinates $(\hat{x}, \hat{y}, \hat{z})$ using the least squares method to achieve an optimized solution.

\begin{figure}[]
	\centering
	\includegraphics[width=1\linewidth]{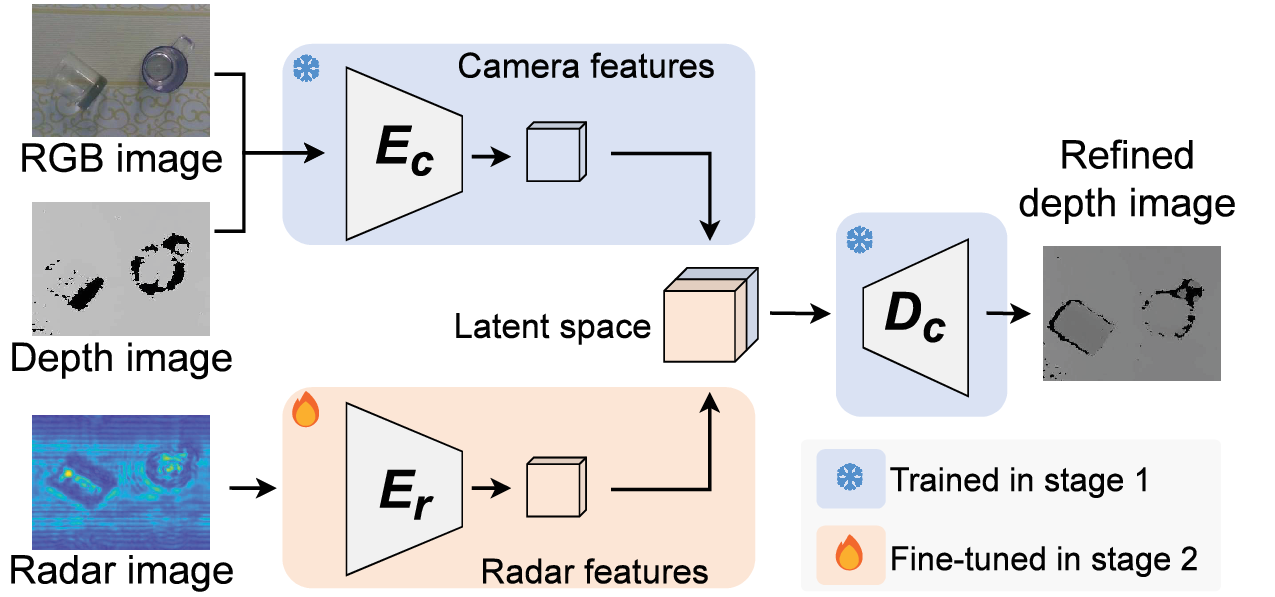}
	\caption{\textbf{Overview of our fusion network architecture.} The network begins with pre-training the camera encoder ($E_c$) and decoder ($D_c$) on a large public RGB-D dataset. Subsequently, the radar encoder ($E_r$) is fine-tuned using our own RGB-D-Radar dataset.}
	\label{Network}
	\vspace{-1em}
\end{figure}

We now model the signal offsets based on $r(t)$ defined in Eq. (\ref{eqn2}). The $l$-th channel error is expressed as the residual delay $\tau_l$ in the phase term, and the received raw data can be written as 
\begin{equation}\label{eqn10}
	\begin{aligned}
		\Tilde{r}(t) = g_l e^{j 2 \pi (f_0+Kt)(\tau + \tau_l)} = \alpha e^{j 2\pi (f_0+Kt) \tau_l} r(t) = \alpha e^{j \phi_l} r(t),
	\end{aligned}
\end{equation}
where $\alpha e^{j \phi_l}$ is the residual complex gain factor. The amplitude $\alpha = e^{j 2 \pi f_0 \tau_l}$ is a constant parameter, while the phase term $\phi_l = 2\pi K\tau_l t =  2\pi f_l t$ is the phase error we need to estimate and compensate. The reference signal $r(t)$ can be generated based on Eq. (\ref{eqn2}) using the estimated $(\hat{x}, \hat{y}, \hat{z})$. By parameter estimation theory \cite{Calib3}, we can convert the estimation of $f_l$  into a frequency estimation problem by multiplying the complex-conjugate reference signal. Mathematically, we have  
\begin{equation}\label{eqn11}
	\begin{aligned}
		\hat{f}_l = \arg  \max_{f} \sum_{l} |W_l(f)|^2, \;\; W_l(f) = \int_0^T \Tilde{r}(t) r^* (t) e^{-j2 \pi ft} dt,
	\end{aligned}
\end{equation}
where $\hat{f}_l$ represents the estimated value of $f_l$. And we can use the FFT to obtain the discrete version of $W_l(f)$ \cite{buhari2018multicarrier}. Finally, the residual complex gain factor can be computed and compensated by substituting $\hat{f}_l$ in Eq. (\ref{eqn10}). Fig. \ref{Calibration} shows the peak profile of a point target along the horizontal direction. We can see that the beamwidth of the point target is significantly improved after signal calibration. Moreover, the signal strength is increased and the quality of SAR images is enhanced. 
A Savitzky–Golay (S-G) filter is further applied to remove DC and high-frequency components from the received radar data \cite{SG}. Additionally, we pre-acquire radar images of the background workspace to perform background cancellation, further enhancing the quality of the radar images.

\subsection{Depth Reconstruction Module} \label{subsec_depth_module}
In the depth reconstruction module, we enhance radar data processing and fuse it with visual imagery using deep learning. To this end, we built our own RDB-D-Radar dataset for transparent objects, comprising 600 data pairs, tailored for depth reconstruction tasks. However, the challenge of training an end-to-end network with this dataset alone arises from its comparatively small size.

\begin{table} [t]
	\caption{The parameters of the CVAE network for transparent depth reconstruction.}
	\centering
	\begin{adjustbox}{max width=0.46 \textwidth}
		\begin{tabularx}{0.6\textwidth} { 
				*{10} {>{\centering\arraybackslash}X }
				>{\centering\arraybackslash}X 
			}
			\toprule[1.3pt]
			& \multicolumn{4}{c}{\textbf{Encoder}} & \textbf{Latent} & \multicolumn{5}{c}{\textbf{Decoder}} \\  
			\cmidrule(r){2-5} \cmidrule(r){6-6} \cmidrule(r){7-11} 
			
			Layer & 1 & 2 & 3 & 4 &  & 1 & 2 & 3 & 4 & 5 \\ \hline
			
			Chan. & 8 & 16 & 32 & 64 & 128 & 64 & 32 & 16 & 4 & 1 \\ 
			
			Kernel & [3,1,1] & [3,2,1] & [3,2,1] & [3,2,1] &  & [3,2,1] & [3,2,1] & [3,2,1] & [3,2,1] & [3,1,1] \\ 
			\bottomrule[1.3pt]
		\end{tabularx}
	\end{adjustbox}
	\vspace{-1em}
	\label{network_arc}
\end{table}

To overcome this constraint, we draw upon extensive, publicly available transparent RGB-D datasets. We propose a unique training strategy that employs these datasets to supplement the depth reconstruction process. Notably, FuseGrasp uses the TransCG dataset, which boasts over 50,000 RGB-D images of more than 50 distinct objects in common indoor settings \cite{transcg}. While these datasets do not specifically consider the extreme lighting environments pertinent to our study, they nonetheless offer valuable image depth cues that enhance our reconstruction capability.

To adequately utilize TransCG dataset, we propose a two-stage training strategy.
In the first stage, we employ a conditional variational autoencoder (CVAE) network to extract camera features. Our objective is to obtain the refined depth image $\lambda_d$, given an initial depth image $\hat{\lambda}_d$ with missing details and the corresponding RGB image. 
We assume that depth images of all transparent objects adhere to a Gaussian distribution, denoted as $P(\Lambda)$. Following the standard CVAE training scheme, we estimate the mean $\mu$ and variance $\sigma$ of $P(\Lambda)$ based on camera features. Subsequently, we sample {from the latent space} and reconstruct the depth map using an decoder.
In this stage, we can already recover a rough depth map only from RGB-D images, but the quality is limited due to the missing radar images. In the second stage, we improve depth quality using supplementary radar images. Since the latent space encapsulates the coarse estimated mean $\mu$ and variance $\sigma$, we employ a fine-tuning strategy to optimize the variance $\sigma$ using the radar images, enabling a more accurate estimation of $P(\Lambda)$. Specifically, we reuse the same decoder from the CVAE and concatenate the radar features with camera features in the latent space. An overview of the fusion network is illustrated in Fig.~\ref{Network}.

\subsubsection{Network Architecture} Table \ref{network_arc} shows the {parameters} of our fusion neural network. The details are given as follows.

\textbf{Camera encoder ($E_c$)}: We use convolution neural networks (CNNs) to encode the high-level latent representation from the public dataset. We stack basic blocks to construct the encoder, where each block consists of a 2D convolution layer, batch normalization, and ReLU activation. The number of channels increases in deeper layers to extract more diverse features. 

\textbf{Camera decoder ($D_c$)}: The decoder has the same block design as the encoder, but with a decreasing number of channels and increasing output tensor sizes in stacked blocks.

\textbf{Radar encoder ($E_r$)}: The radar encoder has a similar design to the camera encoder, but with fewer layers to avoid overfitting to our limited radar images. Additionally, we down-sample the radar images to match the camera feature. The output features of the radar encoder are fused with the latent space extracted in the first stage before being input to the decoder for reconstruction.

\begin{figure}[]
	\centering	
	\subfigure[Radar and RGB images.]{
		\includegraphics[width = 0.6\linewidth]{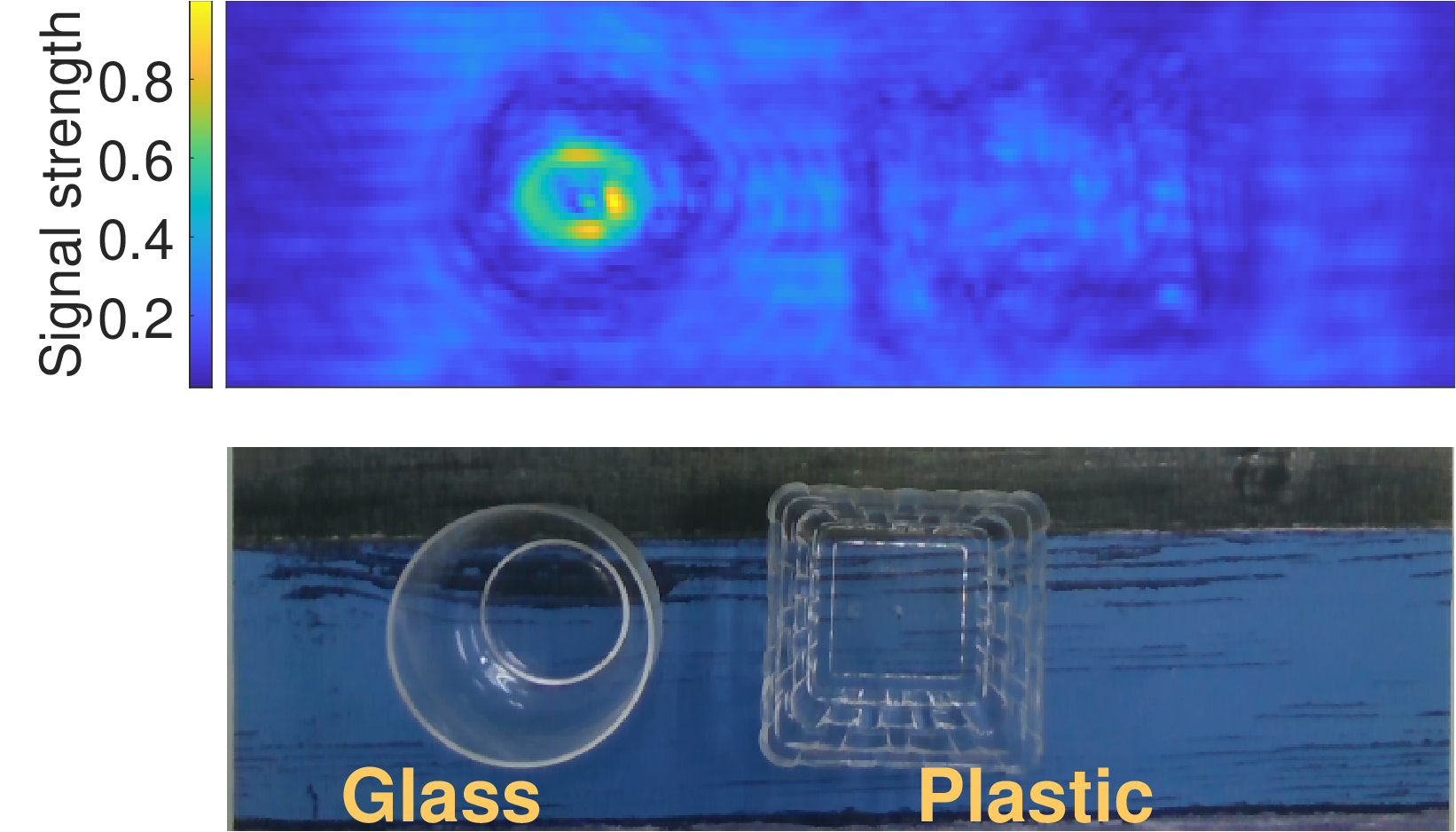}} 
	\label{glassVSplastic_img}
	\subfigure[Signal strengths.]{
		\includegraphics[width = 0.33\linewidth]{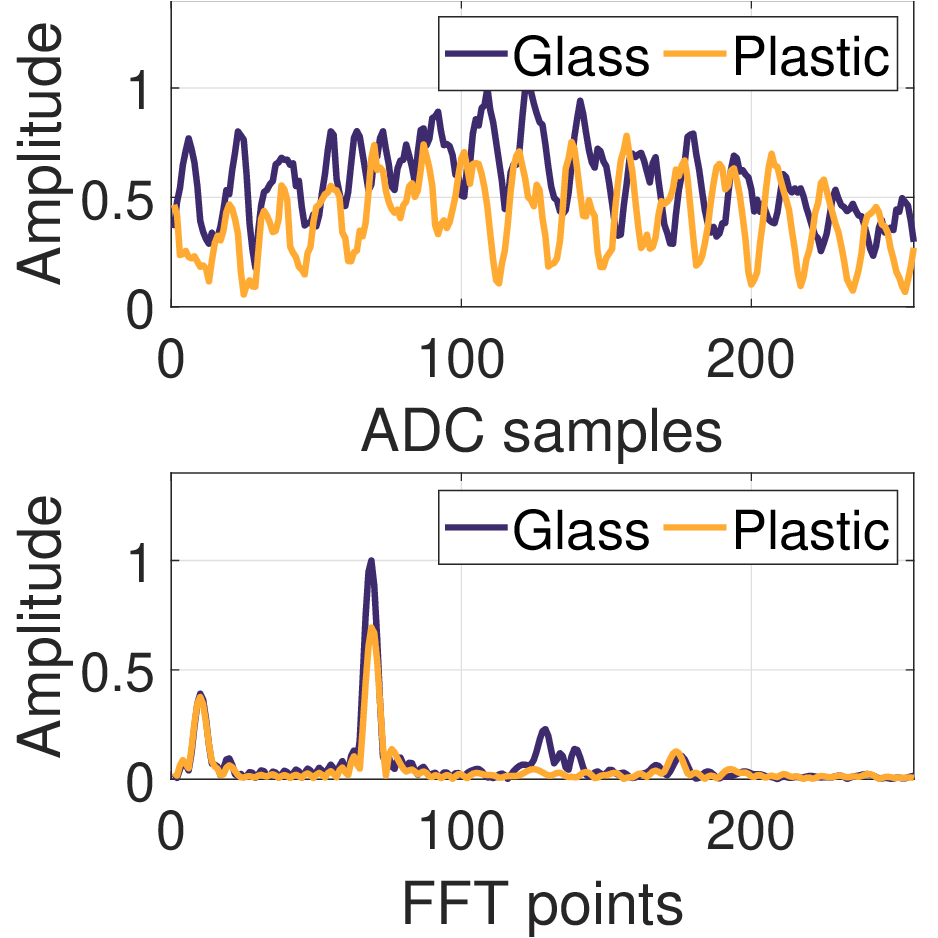}} 
	\label{glassVSplastic_wave}
	
	\caption{\textbf{Material characteristics of two transparent objects.} (a) Radar and RGB images. (b) Received raw signal (top) and the corresponding FFT spectrum (bottom). }	
	\label{MI_radar}
\end{figure}

\subsubsection{Design of Loss Function} To train the network for reconstructing the shape of transparent objects, we carefully design three loss terms, including the depth pixel loss $\mathcal{L}_d$, surface normal loss $\mathcal{L}_{sn}$, and Kulback-Leibler loss $\mathcal{L}_{KL}$. Let the input initial depth image with missing details be $\hat{D} \in \mathbb{R}^{H\times W}$, the output refined depth image be $D \in \mathbb{R}^{H\times W}$, where $H\times W$ denotes the image size. The loss terms are defined as follows:
\begin{equation}\label{eqn12}
	\begin{aligned}
		\mathcal{L}_d &= \sum\nolimits_{p} \lVert D(p) - \hat{D}(p) \rVert^2,
		\\
		\mathcal{L}_{sn} &= \sum\nolimits_{p} \left\{1 -  \cos \Big \langle D_h(p) \times D_w(p), \hat{D}_h(p) \times \hat{D}_w(p) \Big \rangle\right\},
	\end{aligned}
\end{equation}
where $p$ is the pixel index in depth images, $D_w$ and $D_h$ are gradient vectors along width-axis and height-axis of the depth map $D$, respectively. Further, the Kulback-Leibler loss $\mathcal{L}_{KL}$ represents the information divergence, which regularizes the distribution $P(\Lambda)$ to be Gaussian. The overall loss function~is:
\begin{equation}\label{eqn13}
	\begin{aligned}
		\mathcal{L} = \mathcal{L}_d + \alpha \mathcal{L}_{KL} + \beta \mathcal{L}_{sn},
	\end{aligned}
\end{equation}
where $\alpha$ and $\beta$ are weight parameters. When calculating these losses, we regard depth values outside the range $[0.3, 0.6]$ as invalid pixels and exclude them to reduce the impact of outliers.

\subsection{Material Identification Module} \label{Section_MI}
In this module, we exploit the unique scattering coefficients of different materials, which enables precise material characterization through the analysis of mmWave images. Although the scattering coefficient can be influenced by factors such as the incident signal angle and surface texture, it is primarily determined by the material's inherent properties. To validate this, we conducted an experiment featuring a glass and a plastic bowl placed on a table. The results, as shown in Fig. \ref{MI_radar}a, distinctly highlight the different radar signatures of the two materials. We note that the radar image of the glass bowl is significantly brighter than that of the plastic bowl, and the amplitude of the raw received signal is higher, as depicted in Fig. \ref{MI_radar}b. This is consistent with the expectation that glass, being a stronger reflector, will produce a more pronounced radar return than plastic, which is a weaker reflector.

We develop a robust classifier to distinguish between materials. Building upon established image recognition frameworks like YOLO \cite{YOLO}, we map 2-D radar images to corresponding material categories. To ensure that the network learns to identify materials rather than memorizing the shape-material correlation, we included training items that are visually similar yet made of different materials. 
Our evaluation of material recognition performance was conducted using a binary classification YOLO v3 network \cite{yolov3}, which was annotated and trained specifically for this purpose.

\subsection{Grasping Module}
FuseGrasp is designed to efficiently handle the picking of detected transparent objects within a workspace. To accomplish this, we integrate the advanced robot grasping algorithm GR-ConvNet \cite{Grasp_model}, a neural network adept at generating a range of viable 2-D grasp poses. We begin by adjusting the resolution of the provided RGB-D image and radar data to fit our depth map reconstruction module's requirements. Once processed, the enhanced depth image is restored to its original size and presented as input to the grasping network. The network then proposes potential grasp candidates along with their respective 2-D poses. Following this, the robot proceeds to carry out the grasping operation using a parallel-jaw gripper.

\section{Implementations}\label{sec_implement}

\subsection{Hardware Implementations}
\begin{figure}[]
	\centering	
	\includegraphics[width = 0.95\linewidth]{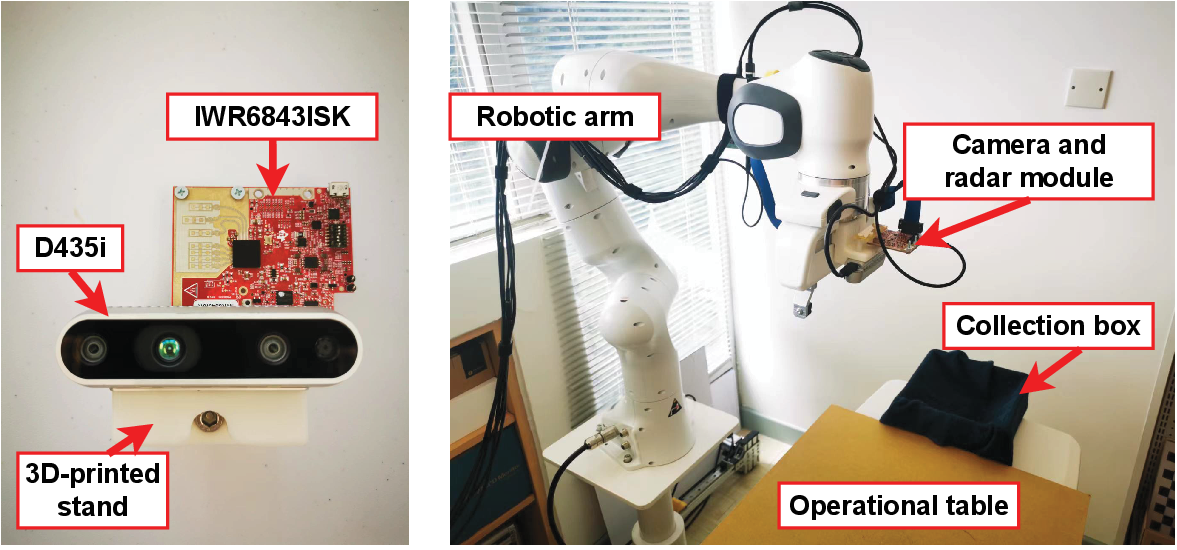} 
	\caption{\emph{\textbf{Left.}} Hardware implementation. \emph{\textbf{Right.}} Setup of experiment.}
	\label{experiment}
	\vspace{-1.5em}
\end{figure}

We implement FuseGrasp using a TI IWR6843ISK mmWave radar, equipped with etched antennas for its four receivers and three transmitters \cite{TI}. Operating at a center frequency of 61.8 GHz and a bandwidth of 3.6 GHz, our setup is optimized to reduce power consumption by using only 2$\times$4 Tx-Rx pairs. 
Additionally, we set the chirp frame period to 10 ms, the sample rate to 4400 kHz, and each frame contains 256 sample points.
This configuration forms an eight-channel linear antenna array that operates in time division multiplexing mode. The mmWave radar unit is mounted alongside an Intel Realsense D435i RGB-D camera \cite{Intel} on a custom-designed 3D-printed stand, creating a unified sensor apparatus for the robot, as illustrated in Fig. \ref{experiment}. A 7-DoF Franka robotic arm \cite{Franka} is employed to maneuver the mmWave radar and grasp the detected transparent objects. The entire system operates on Ubuntu 18.04 and interfaces with the devices through Ethernet connections for robust data transmission and control.

\subsection{Software Implementations}
We use MATLAB for signal processing and employ Python in conjunction with with C++ for controlling the operations of the robotic arm. Within our framework, the Adam optimizer is employed, initiated with a learning rate of $10^{-3}$. A multi-step learning rate scheduler is used to methodically decrease the learning rate at intervals, specifically after every 5 epochs. The training process consisted of two stages: In the first stage, we train the model for 100 epochs with a batch size of 32. In the second stage, we freeze the parameters of the pre-trained model and focus solely on training the radar encoder for 500 epochs with a batch size of 16. For the loss function, we empirically set the coefficients $\alpha = 0.1$ and $\beta = 0.01$. Using an NVIDIA GeForce RTX A6000 GPU, we train on the TransCG datasets for a total of approximately 96 hours.

\section{Evaluation}\label{sec_evaluation}
In this section, we evaluate FuseGrasp using different transparent objects placed on the operational table.
Most of these objects have appeared in our dataset, yet some are unseen before. Moreover, we also prepare different wallpapers to create different background textures. In the experiment, the robotic arm starts from a height of 0.6m to obtain the RGB-D images, as shown in Fig. \ref{experiment}. Then it automatically sweeps over the operational table to generate a uniform linear antenna planar array with 161 $\times$ 84 virtual antenna elements along the horizontal and vertical axes. The 3dB beamwidth of the mmWave radar ensures that the synthetic  planar array can cover the entire workspace. Such a sweeping process takes approximately 30 seconds. We avoid scanning only the local area that may contain the target, because i) the RGB-D camera may fail to locate the transparent object, and ii) scanning multiple local areas may take more time than scanning the whole area. The resolution of captured RGB-D image and radar image are 640 $\times$ 480 and 320 $\times$ 160, respectively. To reduce the computing resource, we resize the RGB-D images to $320 \times 240$ and the radar images to $160 \times 120$ during model training and testing. In the following, we describe the comparison baselines for transparent object radar imaging, shape reconstruction and material identification, then discuss the evaluation results,  respectively.

\subsection{Radar Imaging}
We now report more quantitative evaluations of radar imaging.

\subsubsection{Distance Estimation Accuracy}
\begin{figure}[]
	\centering	
	\includegraphics[width = 0.9\linewidth]{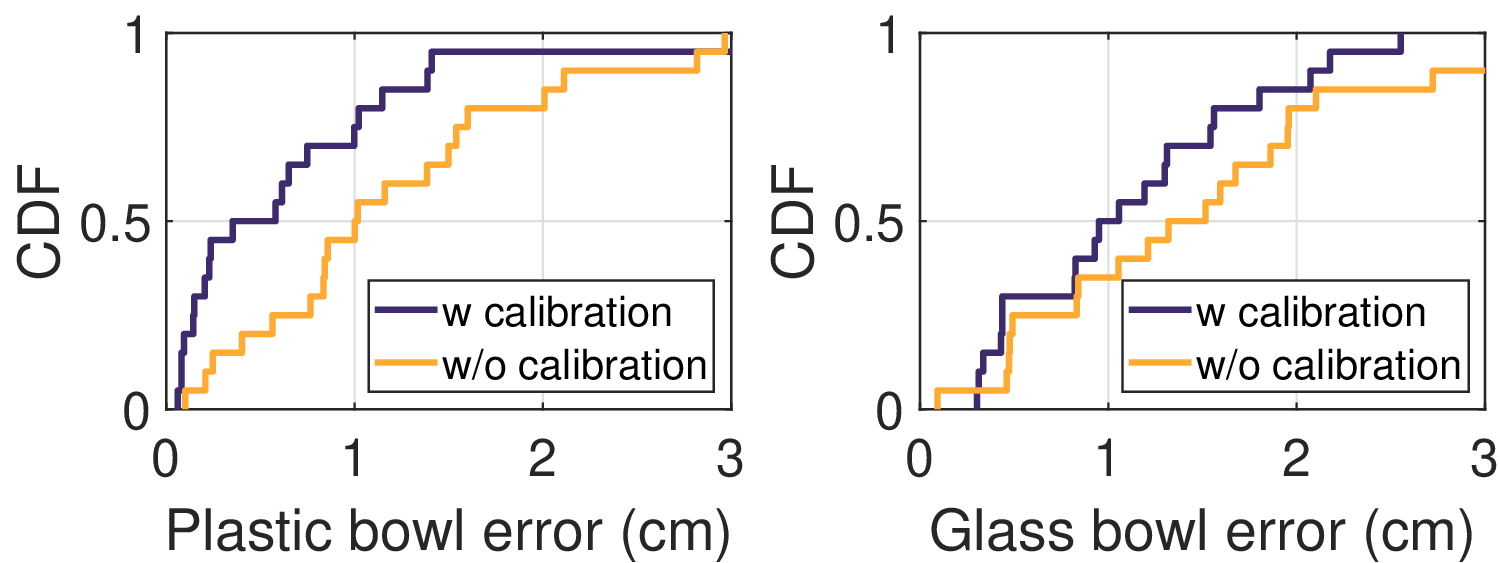} 
	\caption{CDFs of distance estimation errors for two transparent bowls.}	
	\label{radar_acc}
	\vspace{-1.5em}
\end{figure}

One widely-used metric for assessing the quality of radar images is distance estimation accuracy. Therefore, in our experiment, we evaluate planar distance accuracy along the radar scanning direction. We prepare two bowls made of different materials in this experiment, where the plastic bowl is larger than the glass one. Their outer diameters (i.e., the ground truth values in distance estimation) are 12 cm and 10 cm, respectively. The error is defined as the absolute difference between the estimated distance $r_e$ and the actual distance $r_a$, i.e., $|r_e - r_a|$. Fig. \ref{radar_acc} plots the cumulative distribution functions (CDFs) of distance estimation errors for different materials of transparent bowls. After signal calibration, we observe that the median errors are 0.47 cm and 0.98 cm for the plastic bowl and glass bowl, respectively. These errors are consistent with the common understanding that the larger radar cross-section (RCS) of the plastic bowl causes higher reflection of mmWave signals, resulting in smaller localization errors. In contrast, the raw radar images indicate a serious performance degradation: median errors are 0.99 cm and 1.41 cm for the same two cases. These observations confirm that noise and phase error significantly affect the performance of radar imaging, and our calibration algorithm is effective against different materials.

\subsubsection{Imaging Quality}
\begin{figure}[]
	\centering	
	\includegraphics[width = 1\linewidth]{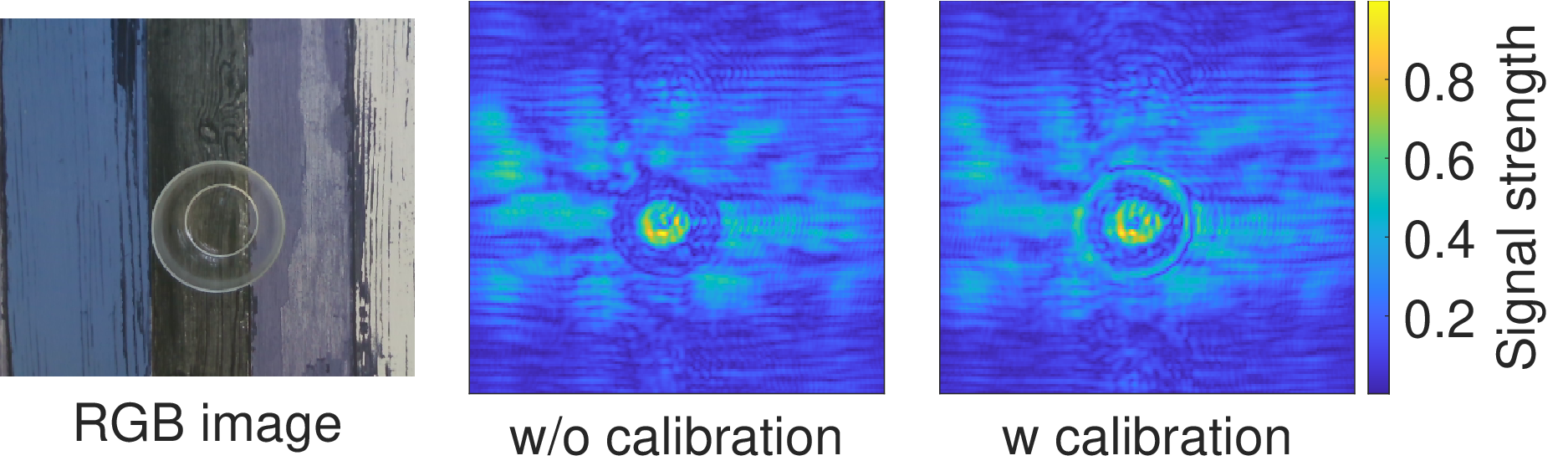} 
	\caption{Radar imaging results of one glass bowl.}
	\label{radar_img}
\end{figure}

To enhance RGB-D image augmentation for transparent objects, it is essential to have high-quality radar images to compensate for depth inaccuracies. Fig.~\ref{radar_img} illustrates the improvement in imaging quality achieved through signal calibration. Notably, the glass bowl's edge definition is sharper after calibration, offering more valuable geometric details. We employ image entropy as the quantifier of imaging quality \cite{SAR_quality}. Image entropy is indicative of the imaging results' focus, where a lower entropy value signifies higher quality. Post-signal calibration, the entropy value decreases from 6.70 to 5.92, evidencing a modest yet noteworthy enhancement in the overall imaging quality of transparent materials due to the calibration algorithm.

\subsection{Depth Completion} \label{subsec_depth_result}
In this section, we undertake a rigorous evaluation of the depth completion efficacy of FuseGrasp. The fidelity of shape reconstruction exerts a significant impact on the precision of robotic arm grasping maneuvers. First, we introduce the evaluation metrics that have been selected for this study. Subsequently, we conduct a comparative analysis of the depth completion performance of FuseGrasp against two established frameworks—ClearGrasp \cite{cleargrasp} and TransCG \cite{transcg}. Finally, we critically evaluate the effects of various factors on the robustness and reliability of FuseGrasp.

\subsubsection{Evaluation Metrics}
We use a set of widely recognized metrics for the depth completion of transparent objects. Unless stated otherwise, all metrics are calculated solely within the transparent regions as delineated by the corresponding transparency masks.

\textbf{RMSE}: The root mean squared error between estimated depths and groundtruth depths.

\textbf{MAE}: The mean absolute error between estimated depths and groundtruth depths.

\textbf{Threshold} $\delta$: The percentage of pixels with the predicted depth $\hat{d}$ satisfying $ \max(\frac{\hat{d}}{d}, \frac{d}{\hat{d}}) < \delta $, where $\hat{d}$ and $d$ are corresponding pixel-wise depths in estimated depth image $\hat{D}$ and groundtruth depth image $D$ respectively, and $\delta$ is set to 1.05, 1.10 and 1.25 \cite{CVPR_ref1}.

We also use the estimated errors in depth distance as depth completion evaluation metrics when using size-fixed transparent cylinder and cuboid.

\begin{table}[]
	\caption{Comparison with four baselines on self-built dataset.}
	\centering
	\begin{threeparttable}
		\begin{adjustbox}{max width=0.75\textwidth}
			\begin{tabular}{c|ccccc}
				\hline
				\multicolumn{1}{c|}{\multirow{3}{*}{\textbf{Methods}}} & \multicolumn{5}{c}{\textbf{Depth Completion Metrics}} \\ \cline{2-6} 
				\multicolumn{1}{c|}{} & RMSE $ \downarrow $ & MAE $ \downarrow $ &$  \delta_{1.05} $ $ \uparrow $ & $ \delta_{1.10} $ $ \uparrow $ & $ \delta_{1.25} $ $ \uparrow $ \\
				\multicolumn{1}{c|}{} & (m) & (m) & (\%) & (\%) & (\%) \\ \hline
				ClearGrasp & 0.220 & 0.171 & 7.593 & 28.20 & 74.01 \\ 
				TransCG & 0.109 & 0.072 & 43.24 & 60.78 & 77.65 \\
				Direct fusion & 0.119 & 0.061 & 35.43 & 62.58 & 80.12 \\  
				Our (w/o cal.) & {0.091} & {0.044} & {44.85} & {73.01} & {90.19} \\ \hline
				\rowcolor{lightgray} Our (with cal.) & \textbf{0.058} & \textbf{0.041} & \textbf{45.78} & \textbf{77.91} & \textbf{92.40} \\ \hline
			\end{tabular}
		\end{adjustbox}
		\begin{tablenotes}
			\item \textbf{Note.} $ \downarrow $ means lower is better, and $ \uparrow $ means higher is better. The best result in each column is \textbf{bold}.
		\end{tablenotes}
	\end{threeparttable}
	\label{total_result}
	\vspace{-1em}
\end{table}

\subsubsection{Overall Performance}
We compare FuseGrasp with four baselines using our self-built dataset. ClearGrasp is the first framework using deep learning with synthetic training data to enhance the depth estimation of transparent objects \cite{cleargrasp}. Meanwhile, TransCG employs a U-Net architecture to undertake the transparent object depth completion task \cite{transcg}. 
{Moreover, we designed and implemented an additional baseline: a direct fusion network. This was done to demonstrate the necessity of developing FuseGrasp. Specifically, the direct fusion baseline uses an Encoder-to-Decoder network similar to DLoc \cite{dloc}, a wireless localization method that fuses 2-D signal heatmaps from different sensors.}
To ensure fair comparisons, we trained the ClearGrasp and TransCG models on our dataset using publicly available source codes and the recommended optimal hyper-parameter settings provided by the authors. We also trained the direct fusion network on a public RGB-D dataset \cite{transcg} and then fine-tuned it on our self-built dataset. Furthermore, to demonstrate the benefits of radar signal calibration (see Sec. \ref{sec:imgage}), we fine-tuned FuseGrasp using two versions of our self-built dataset: one without radar signal calibration and one with it.

\begin{figure*}[]
	\centering
	\includegraphics[width= 0.85\textwidth]{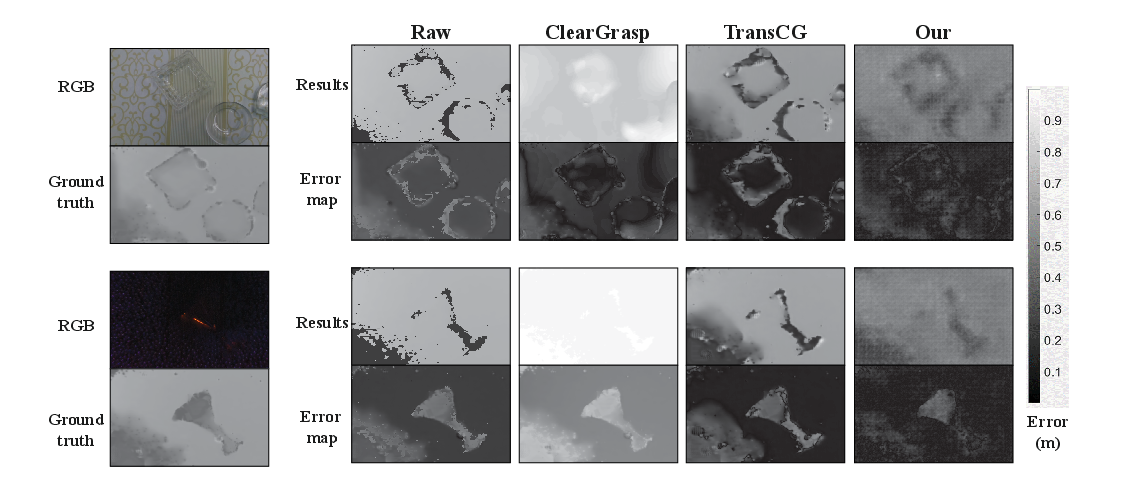}
	\caption{Visual comparisons of reconstructed depth images and error maps using self-built dataset: normal group (top) and dim group (bottom). In the error maps, darker shades correspond to regions of reduced error.}
	\label{Visualizations}
	\vspace{-1em}
\end{figure*}

Fig.~\ref{Visualizations} visually compares the reconstructed depth images and error maps of FuseGrasp with benchmark methods. It shows results from ClearGrasp, TransCG, and our FuseGrasp using grayscale images for two data groups. Darker regions in the error maps indicate lower errors. Due to the limited sample size of our self-built dataset, transparent object boundaries appear blurry in the results of FuseGrasp. However, the quantitative results in Table \ref{total_result} show that FuseGrasp yields lower errors compared to the two baselines. We observe that due to the asymmetry in different modalities, directly fusing RGB-D and radar data results in marginal improvements or may even degrade performance in certain metrics. In contrast, FuseGrasp excels in all evaluated metrics, and radar signal calibration further enhances its performance.

\subsubsection{Impact of Two-stage Training}
We validate the effectiveness of the two-stage training strategy by comparing two scenarios: first, training the entire neural network using only our self-built dataset; second, following the two-stage strategy by freezing the CVAE component and only fine-tuning the radar encoder. Table~\ref{full} shows that the fine-tuning approach achieves better performance on all metrics on the testing set. This improvement arises from the fact that our self-built dataset is too small to effectively train the entire end-to-end network. Therefore, the two-stage fine-tuning strategy proves to be more suitable for our task, given the limited size of our custom dataset.

\begin{table}[]
	\caption{Ablation studies of FuseGrasp.}
	\centering
	\begin{adjustbox}{max width=0.48\textwidth}
		\begin{tabular}{c|ccccc}
			\hline
			\multicolumn{1}{c|}{\multirow{3}{*}{\textbf{Methods}}} & \multicolumn{5}{c}{\textbf{Depth Completion Metrics}} \\ \cline{2-6} 
			\multicolumn{1}{c|}{} & RMSE $ \downarrow $ & MAE $ \downarrow $ &$  \delta_{1.05} $ $ \uparrow $ & $ \delta_{1.10} $ $ \uparrow $ & $ \delta_{1.25} $ $ \uparrow $ \\
			\multicolumn{1}{c|}{} & (m) & (m) & (\%) & (\%) & (\%) \\ \hline
			\multicolumn{1}{l|}{- w/o Fine-tune} & 0.160 & 0.144 & 4.891 & 10.74 & 43.01\\ 
			\multicolumn{1}{l|}{- w/o Radar Images} & 0.173 & 0.151 & 4.358 & 9.371 & 39.57 \\ 
			\multicolumn{1}{l|}{- w/o RGB Images} & 0.127 & 0.094 & 8.825 & 13.96 & 51.73 \\ 
			\multicolumn{1}{l|}{- w/o Depth Images} & 0.185 & 0.162 & 4.144 & 8.771 & 37.73 \\
			\rowcolor{lightgray}\multicolumn{1}{l|}{FuseGrasp} & \textbf{0.058} & \textbf{0.041} & \textbf{45.78} & \textbf{77.91} & \textbf{92.40}\\ \hline
		\end{tabular}
	\end{adjustbox}
	\vspace{-1.5em}
	\label{full}
\end{table}

\subsubsection{Impact of Radar Image}
Radar images also play an important role in the depth reconstruction. To verify that our model benefits from radar images, we train a model using only RGB-D images. Specifically, we achieve this by removing the radar encoder and training the network only on RGB-D inputs. Table \ref{full} shows that the camera-only model performs worse than FuseGrasp. This observation can be attributed to the inclusion of both normal and dim lighting conditions in the testing data. Under dim lighting, the camera data quality deteriorates, leading to less effective reconstruction in these challenging scenarios when radar data is not available as complementary information.

\subsubsection{Impact of RGB and Depth Images}
RGB and depth images both play important roles in the depth reconstruction process. To verify that our model effectively integrates and benefits from these two modalities, we train two separate models: one uses only RGB-Radar images, and the other only Depth-Radar images. Specifically, we modify the number of input channels of the RGB-D encoder and train the network with RGB-Radar and Depth-Radar data pairs as inputs respectively.
The testing data includes both normal and dim lighting conditions. As shown in Table \ref{full}, the performance of both models is inferior to that of FuseGrasp. Notably, the Depth-Radar model outperforms the RGB-Radar model. This observation can attributed to the depth sensor's ability to capture the shape of transparent targets, even under low-light conditions, despite the presence of missing or inaccurate points.

\subsubsection{Impact of Target Height}
The height of a transparent target can affect the depth completion performance of FuseGrasp, as taller targets closer to antennas will reflect more mmWave signal. We use different transparent solid cylinders and cuboids to evaluate the effect of target height on the depth reconstruction results. Fig. \ref{relationship_envs}a shows that as height increases from 5cm to 15cm, the median depth distance error decreases by 1.11 cm on average. This indicates that more mmWave signals are reflected by taller targets. The cuboid performs better than the cylinder, as it has a larger cross-section. The attenuation of mmWave signals as they penetrate the target is negligible for small targets in daily life. The increased height does not cause a significant increase in error, because the RGB-D images also contribute to the shape reconstruction and compensate for the radar images. This confirms the robustness of FuseGrasp in adapting to height variations.

\subsubsection{Impact of Background Texture}
As we discussed in Section \ref{sec_introduction}, traditional camera-based depth reconstruction algorithms are sensitive to the background texture of the workspace. Here, we verify whether FuseGrasp is robust to different background textures. We place the transparent targets on three hard cardboard with different wall stickers, which can be classified as non-textured, simple-textured, and complex-textured backgrounds. The evaluation results are shown in Fig. \ref{relationship_envs}b. For the same target, the median depth error changes little for different backgrounds. This confirms that FuseGrasp is robust to different background textures. We note that the median error exhibits a modest average fluctuation of 0.87 cm when different textures are introduced, which could be attributed, in part, to FuseGrasp's reliance on camera images for depth reconstruction. As previously discussed, the radar imagery serves as complementary data to augment the target shape delineation derived from the RGB-D image. Fortunately, FuseGrasp demonstrates robustness against such variability, maintaining reliable performance despite the influence of differing textures.

\subsubsection{Impact of Lighting Condition}
\begin{figure*}[htbp]
	\centering
	\subfigure[Target height.]{
		\includegraphics[width = 0.18\linewidth]{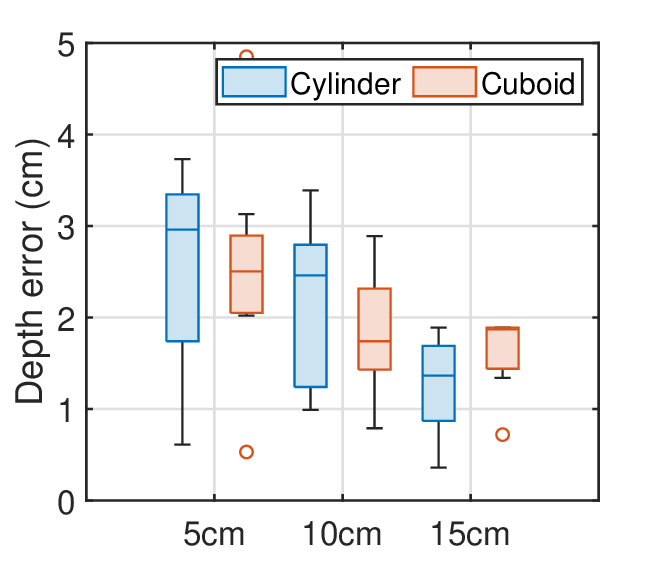} }
	\label{envs_height}
	\subfigure[Background texture.]{
		\includegraphics[width = 0.18\linewidth]{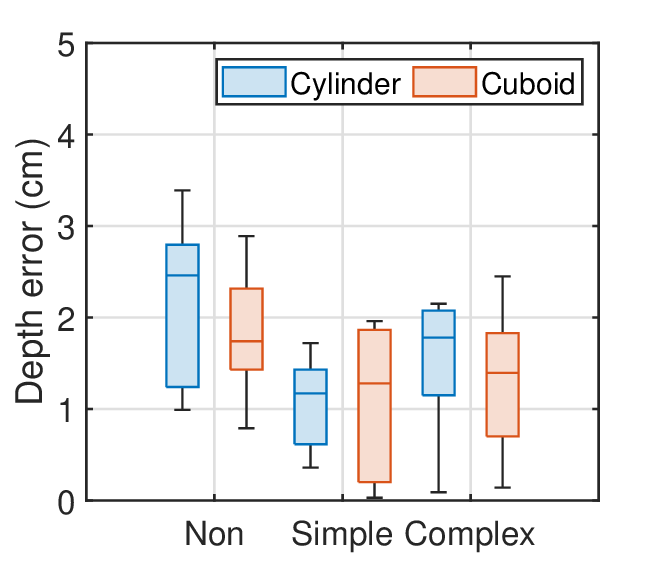}} 
	\label{envs_background}
	\subfigure[Lighting conditions.]{
		\includegraphics[width = 0.18\linewidth]{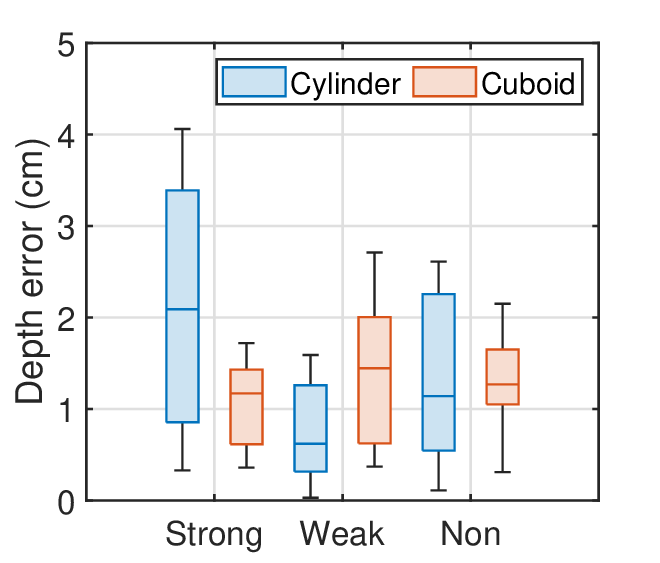}} 
	\label{envs_lighting}
	\subfigure[Capturing elevations.]{
		\includegraphics[width = 0.18\linewidth]{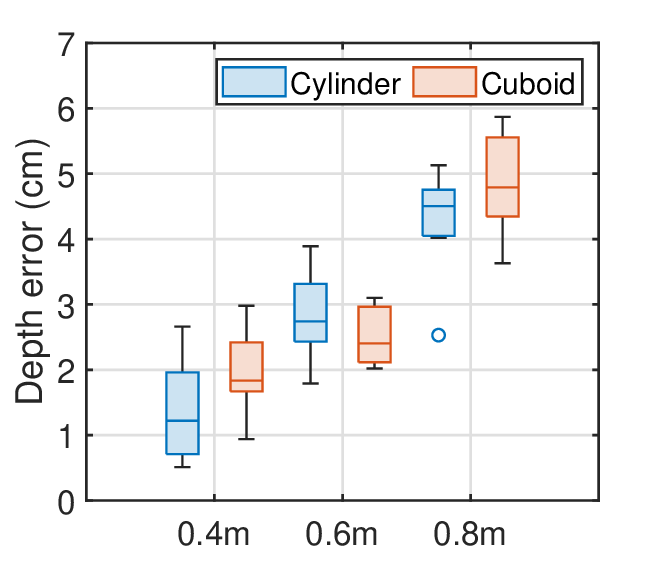} }
	\label{envs_capturing}
	\subfigure[Material.]{
		\includegraphics[width = 0.18\linewidth]{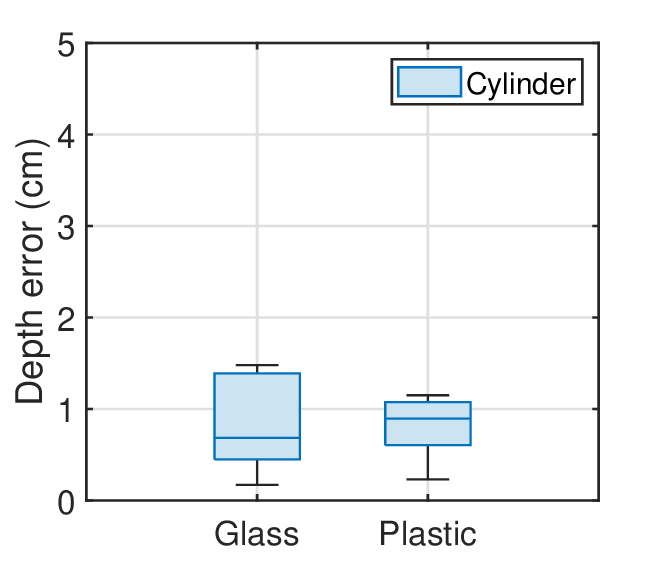} }
	\label{envs_material}       
	\caption{The impacts of various factors on depth estimation errors.}	
	\label{relationship_envs}
\end{figure*}

The quality of camera images deteriorates as the intensity of the lighting changes, hence we also test whether FuseGrasp can reconstruct depth images clearly under different lighting conditions. We place transparent objects on the same workspace at different hours of the day, and measure the error in depth, with the results shown in Fig. \ref{relationship_envs}c. We note that the median depth distance errors for the cylinder are 2.09 cm, 0.62 cm, and 1.14 cm, respectively, for different lighting conditions. Other objects follow a similar trend but appear less sensitive to lighting, and the average change is less than 0.5 cm. It is evident that FuseGrasp can deliver a decent accuracy despite the changes in illumination.

\subsubsection{Impact of Capturing Elevations.}
The robotic data capturing elevations can impact the depth completion performance of FuseGrasp since the self-built RGB-D-Radar dataset is collected at a fixed capturing elevation. We use the same transparent solid cylinder and cuboid to evaluate how robotic data capturing elevations affect the depth reconstruction results. As presented in Fig. \ref{relationship_envs}d, when the robotic data capture height is elevated from 0.4 m to 0.8 m, the median depth distance error increases by 3.12 cm on average. This observation can be attributed to the fact that, at a higher height, the observed size of the target is smaller, and less power of mmWave signals is reflected. We note that the corresponding relative depth error shows only a modest increase, rising from 5.09\% to 6.64\%. In practice, while the robotic capturing height in real-world deployments may differ from those used during training data collection, the overall shape of the transparent target remains largely consistent. Additionally, the RGB-D images serve as a complementary input to the shape reconstruction process, effectively compensating for potential inaccuracies in the radar data. As a result, variations in capturing heights do not cause a significant increase in depth reconstruction error. This outcome highlights the robustness of FuseGrasp in handling height variations effectively.

\subsubsection{Impact of Material}
The material of transparent objects also affects the depth completion performance. To investigate the effects of different materials, we use two same-size cylinders: one plastic and one glass. Fig. \ref{relationship_envs}e shows that, for the plastic  cylinder, more than 75\% of the depth error is smaller than 1.07 cm; although the glass cylinder has slightly higher errors in percentile, the median error in depth is below 0.69 cm. We deduce that the difference in reflectivity (glass is a stronger reflector than plastic) may have caused the discrepancy in errors, as the stronger reflector produces higher quality radar images.

\subsubsection{Impact of Robotic Arm Precision Levels}
\begin{figure}[]
	\centering
	\subfigure[Radar images of one glass bowl.]{
		\includegraphics[width = 0.4\linewidth]{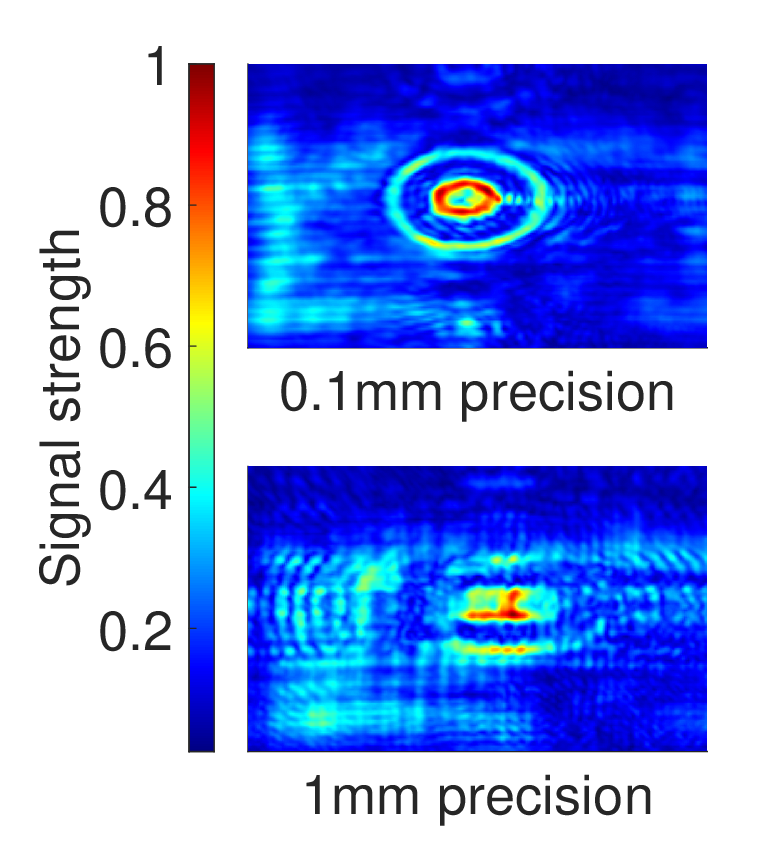} }
	\label{bowl_precision}
	\hfill
	\subfigure[The impacts of precision on depth estimation errors.]{
		\includegraphics[width = 0.5\linewidth]{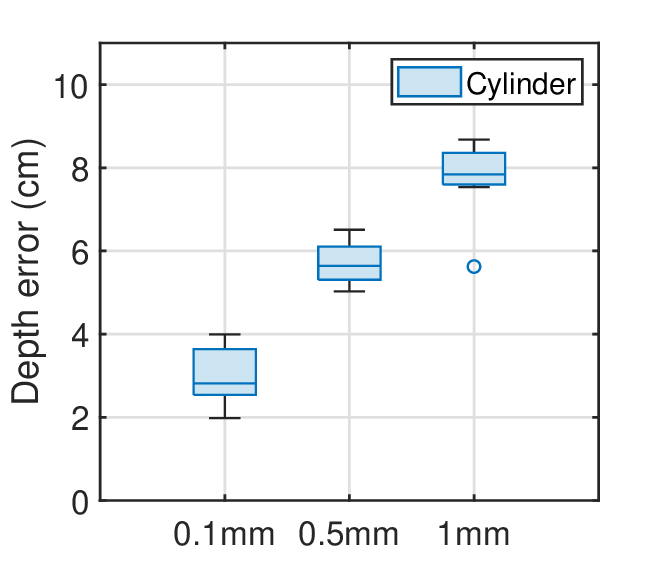}}
	\label{box_precision}
	\caption{The impacts of robot precision levels on radar images and depth estimation errors.}	
	\vspace{-1em}
	\label{radar_precision}
\end{figure}

The quality of radar images depends on the precise movement of the robotic arm. Thus, we further test whether FuseGrasp can clearly reconstruct depth images under varying robotic precision levels. As we do not have access to other robotic arms, we use synthetic data to evaluate the impact of robotic precision on the performance of FuseGrasp. Positional noise was deliberately introduced into the collected data to simulate the behavior of robotic arms with lower precision.
Fig. \ref{radar_precision}a shows radar images of one glass bowl at two different radar precision levels. It can be observed that the circle profile, which is clearly distinguishable at a precision of 0.1 mm (i.e., the precision of Franka robot), becomes blurred when the precision drops to 1 mm.  Given that our mmWave radar has a center frequency of 61.8 GHz, the half-wavelength is around 2.5 mm. As robot precision decreases, the mismatch between adjacent antenna arrays becomes more prominent.
Subsequently, we place a transparent cylinder of a fixed size on the workspace and measure the depth estimation error, presenting the results in Fig. \ref{radar_precision}b. Notably, with robotic precision being 0.1 mm, 0.5 mm and 1 mm, the median depth estimation errors of the cylinder are measured as 2.82 cm, 5.64 cm and 7.85 cm respectively. It is evident that the accuracy of FuseGrasp is influenced by the precision of the robotic arm. However, most industry-grade robotic arms achieve a precision of 0.1 mm or better, ensuring that precision errors have a negligible impact on the performance of FuseGrasp in real-world applications.

\begin{figure}[]
	\centering	
	\subfigure[FuseGrasp.]{
		\includegraphics[width = 0.45\linewidth]{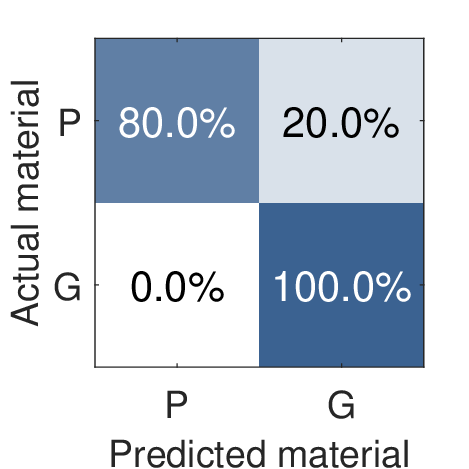}}
	\label{mi_radar}
	\subfigure[Camera-based.]{
		\includegraphics[width = 0.45\linewidth]{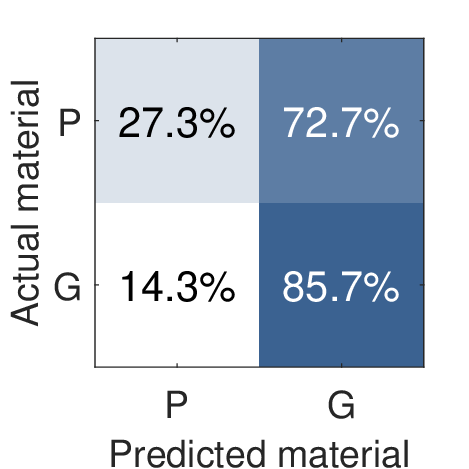}}
	\label{mi_camera}
	\caption{\textbf{Confusion matrices of material identification.} P and G stands for plastic and glass objects, respectively.}	
	\label{Material_Matrix}
	\vspace{-1em}
\end{figure}

\subsection{Material Identification}
We now present the material identification performance of FuseGrasp, and further compare it with a camera-based identification algorithm. We prepare two common transparent materials: glass and plastic. We use confusion matrices in Fig. \ref{Material_Matrix} to show the identification performance of different methods. We observe that the radar-based solution used in FuseGrasp is more robust than the vision-based methods. The overall accuracy for identifying the material (glass or plastic) is 90.0\%, {substantially higher than the camera-based methods (56.5\%)}. Specifically, both radar and camera can identify glass material correctly, although the camera is slightly less accurate. For plastic material, the radar-based solution is much better than its camera counterpart. One reasonable explanation is that the glass items are mostly heavy glass, which makes them easier to see visually. In contrast, the plastic items have different thicknesses, making it harder for camera to identify.

\begin{table}[]
	\caption{\textbf{Performance of real-world robotic arm experiments.} 
		Scene 1 to 4 represent strong sunlight, weak sunlight, indoor light, and no light, respectively.
		We randomly place 3 testing objects in each experiment.}
	\centering
	\begin{adjustbox}{max width=1.3\textwidth}
		\begin{tabular}{|ccc|c|c|}
			\hline
			& & &\multicolumn{1}{c|}{Camera-only}& \multicolumn{1}{c|}{FuseGrasp}\\ \hline
			\# Scenes & \multicolumn{1}{|c}{\# Expr.} & \multicolumn{1}{|c|}{\# Objects} & \# Grasped & \# Grasped \\ \hline
			\multirow{4}{*}{\makecell{Scene 1\\(Strong sunlight)}} 
			& \multicolumn{1}{|c|}{Expr. 1}&3 &0 &3 \\ 
			& \multicolumn{1}{|c|}{Expr. 2}&3 &1 &3 \\
			& \multicolumn{1}{|c|}{Expr. 3}&3 &1 &3 \\
			& \multicolumn{1}{|c|}{Expr. 4}&3 &0 &2 \\ \hline
			
			\multirow{4}{*}{\makecell{Scene 2\\(Weak sunlight)}} 
			& \multicolumn{1}{|c|}{Expr. 5}&3 &2 &3 \\ 
			& \multicolumn{1}{|c|}{Expr. 6}&3 &1 &3 \\
			& \multicolumn{1}{|c|}{Expr. 7}&3 &2 &3 \\
			& \multicolumn{1}{|c|}{Expr. 8}&3 &2 &3 \\ \hline
			
			\multirow{4}{*}{\makecell{Scene 3\\(Indoor light)}} 
			& \multicolumn{1}{|c|}{Expr. 9}&3 &3 &3 \\ 
			& \multicolumn{1}{|c|}{Expr. 10}&3 &3 &3 \\
			& \multicolumn{1}{|c|}{Expr. 11}&3 &2 &3 \\
			& \multicolumn{1}{|c|}{Expr. 12}&3 &3 &3 \\ \hline
			
			\multirow{4}{*}{\makecell{Scene 4\\(No light)}} 
			& \multicolumn{1}{|c|}{Expr. 13}&3 &1 &3 \\ 
			& \multicolumn{1}{|c|}{Expr. 14}&3 &0 &2 \\
			& \multicolumn{1}{|c|}{Expr. 15}&3 &1 &3 \\
			& \multicolumn{1}{|c|}{Expr. 16}&3 &1 &2 \\ \hline
			
			\rowcolor{lightgray}Total & \multicolumn{1}{|c|}{16} & \multicolumn{1}{c|}{48} & \multicolumn{1}{c|}{23} & \multicolumn{1}{c|}{45} \\ \hline
		\end{tabular}
	\end{adjustbox}
	\vspace{-1em}
	\label{real_results}
\end{table}

\subsection{Real-world Robot Experiments} \label{subsec_realrobot}
In this section, we evaluate the performance of FuseGrasp in real-world robotic arm grasping experiments under different lighting conditions. We do not re-train our FuseGrasp due to the limited number of specimens. Instead, we check whether the reconstructed object depth map based on our self-built dataset can generalize to these scenarios.

To demonstrate that FuseGrasp can detect multi-material transparent objects, we selected 9 objects for multi-target retrieval experiments. Three of these objects are unseen during the training phase, and the other 6 objects have appeared in our self-built dataset. We also set different grasping forces for different types of material: we use a stronger force for heavy glass objects, and a weaker force for fragile plastic objects. We perform the grasping experiments at different times of the day to verify the robustness of FuseGrasp under varying light conditions. We conduct 16 experiments under 4 typical scenarios: scene 1 to 4 represent strong sunlight, weak sunlight, indoor light, and no light, respectively. In each scenario, we use the same texture background for a fair comparison. In each experiment, we randomly place 3 test objects near the center of the workspace, and the robotic arm automatically search and grasp them. We repeat each grasping experiment 3 times and also report the result of material identification. 
We use the following two performance metrics: {the grasp completion success rate  defined as $\frac{\text{\# grasped}}{\text{\# objects}}$, and the action efficiency defined as $\frac{\text{\# grasped}}{\text{\# attempts}}$ \cite{zeng2018learning}.}
Each single experiment is deemed successful if the robot detects the objects and then grasps and places them into the collection box.

Table \ref{real_results} reports the experiment results. We can see that FuseGrasp outperforms camera-only grasping algorithms under various lighting conditions. An interesting observation is that when turning on the light at night, the performance of camera-only solution and FuseGrasp are almost close. This result can be attributed to the indoor lighting is softer and there is less shadow interference. However, the performance of camera-only solutions drops significantly in sunlight conditions, where strong light can severely affect the optical and depth sensor data.
The experimental outcomes corroborate the effectiveness of FuseGrasp in manipulating transparent objects composed of diverse materials. Notably, it significantly enhances the grasp completion success rate, amplifying it from 47.9\% to an impressive 93.8\%, even when confronted with unseen objects. Additionally, our empirical observations reveal that while the camera-only solution necessitated 40 attempts to successfully grasp 23 objects, FuseGrasp required only 66 attempts to secure 45 objects successfully. Consequently, FuseGrasp elevates the action efficiency from a modest 57.5\% to an admirable 68.2\%.

\section{Related Work}\label{sec_works}
\subsection{Transparent Depth Completion}
Transparent object depth reconstruction is an active research area employing  various devices such as optical cameras and depth sensors. Some works also introduce additional sensors to complete the depth maps. Existing works can be categorized based on the observation views into multi-view and single-view depth reconstruction.

Multi-view depth reconstruction methods often rely on Neural Radiance Fields (NeRF), which can accurately reconstruct the scene geometry. For instance, Dex-NeRF is designed for depth map reconstruction and transparent object grasping in complex scenes\cite{dex-nerf}, while Evo-NeRF is suitable for sequential transparent object grasping \cite{evo-nerf}. 
Other works like StereoPose, which uses stereo images for transparent object pose and depth estimation, alsorely on multi-view observation \cite{stereopose}.
In single-view depth reconstruction, the first algorithm, ClearGrasp, uses a neural network on synthetic RGB-D datasets to reconstruct depth maps of transparent objects \cite{cleargrasp}. Subsequently, researchers use realistic data instead of synthetic data to train a more robust model. Among these, TransCG, a model employing a U-Net architecture to perform the transparent object depth completion task, has achieved a best performance \cite{transcg}. Some fusion solutions are included in single-view parts. Polarized-CNN combine a polarization camera with an RGB-D camera to search and grasp the transparent targets \cite{kalra2020deep}. TaTa integrates the camera with tactile sensors and can even grasp tiny glass fragments after exploration by a robotic arm \cite{visual_tactile}.

A common limitation of these works is that they almost rely on vision, which may degrade in challenging conditions such as dim light, resulting in lower grasp accuracy. For these additional sensors, the fusion algorithm does not naturally integrate with visual information. To address this, we propose using mmWave radar as a supplementary sensor to provide complementary information for reconstructing the transparent object shape from RGB-D images.

\subsection{Radar Sensing and Imaging}
Radar technology, widely applied in daily life, can be classified into three main types: array signal processing for object sensing \cite{mobi2sense, chen2021movi, khan2022estimating}, point cloud generation for object detection \cite{mmMesh, MilliPoint}, and SAR for shape imaging \cite{zheng2021siwa, UWBMap, MIMO-SAR}.

Array signal processing, based on antenna arrays, is extensively used for radar-based sensing and tracking. For example, Mtrack is an indoor human sensing system that uses beamforming technology to localize multiple moving humans \cite{Mtrack}. Point cloud radars are popular for autonomous driving. MILLIPOINT uses point clouds to detect surrounding objects and enhance driving safety \cite{MilliPoint}. RCVNet employed a feature fusion approach, integrating radar point clouds with visual images, to enable the accurate detection of bird damage in power tower areas \cite{ref2_gao2023rcvnet}.

For SAR imaging applications, UWBMap uses a multi-radar system for indoor mapping and floor plan construction, and it can operate under smoking conditions\cite{UWBMap}. SiWa is another radar imaging work that uses a network to generate SAR images to detect pipes and wires in concrete walls \cite{zheng2021siwa}. MIMO-SAR uses stepping motors to control radars’ point-to-point scanning, and it has achieved high resolution SAR imaging for security checking \cite{MIMO-SAR}.

These three methods have their own advantages and limitations in radar perception. Array signal processing can only locate the target’s position without obtaining its shape details. The sparsity of radar point clouds makes shape reconstruction challenging. SAR imaging can provide geometric shape information to supplement the missing depth data. Therefore, we uses SAR technology to obtain radar images of transparent object, and signal calibration is necessary to improve the radar imaging performance.

\section{Limitation and Future Work}\label{sec_discussion}
\textbf{Search More Effectively.}
The current version of FuseGrasp has a non-negligible searching time that may limit its practical applications. To reduce the searching time, we have optimized the scanning trajectory and used adjacent antennas to collect more receive data simultaneously. However, this trade-off may affect the radar imaging resolution and depth reconstruction accuracy of the system. A possible solution for future work is to apply signal sparse reconstruction techniques to recover the original signal from a small amount of echo data within the same robot workspace.

\textbf{Grasping Policy.}
In Section \ref{subsec_realrobot}, we present that the action efficiency of FuseGrasp is 68.2\%, lower than that achieved by TransCG (80.4\%). This decrease in grasp efficiency for FuseGrasp can be attributed to the unsatisfactory grasping policy. Our chosen grasping network, GR-ConvNet, outputs information related to the robotic arm's coordinates, angle, and hand width within FuseGrasp. However, these 2-D grasping information may not capture the most suitable position as effectively as a 6-DoF pose. Additionally, the 2-D depth image output of FuseGrasp may sacrifice spatial details compared to point cloud data. To overcome these limitations, we plan to construct 3-D point cloud data for training. This will provide more comprehensive spatial information and improve the grasp efficiency of FuseGrasp. Moreover, we aim to develop a 6-DoF pose grasping network, which will enhance FuseGrasp's capability to handle complex scenarios and further improve the grasp efficiency. 

\textbf{Synthetic Dataset.}
As discussed in Section \ref{subsec_depth_result}, the limited size of our self-built dataset can be a limiting factor for the performance of FuseGrasp, underscoring the need for dataset augmentation. Given that collecting realistic radar data would be excessively time-consuming, we plan to follow the Sim2Real approach by synthesizing corresponding radar and RGB-D images in a virtual environment. As data generation occurs entirely under simulated conditions, creating substantial datasets within a limited time is feasible. One of the main challenges of generating synthetic data for FuseGrasp training lie in the lack of radar signal simulation tools. The synthetic dataset will consist of aligned RGB-D-Radar data, and the generated radar signal should respond appropriately to different materials. To overcome this challenge, we plan to design a MATLAB-based radar simulation toolbox and measure realistic data to set an index of refraction that matches the physical model. We also plan to introduce background noise into the dataset to enhance the robustness of model training.

\textbf{Mobile Platform Deployment.} Fusegrasp can be directly adapted to mobile platforms equipped with robotic arms, commonly referred to as mobile manipulators in the literature. The hardware components of FuseGrasp, including the radar and RGB-D camera modules, consume just 1.6W during continuous operation. Furthermore, the network model requires approximately 400MB of GPU memory during inference and can run efficiently on a laptop. These characteristics ensure compatibility with the computational and power constraints of mobile platforms.

\vspace{-1em}
\section{Conclusion}\label{sec_conclusion}
Transparent objects are prevalent in everyday environments, but their distinct physical properties pose significant challenges for camera-guided robotic arms. In this paper, we propose FuseGrasp, the first radar-camera fusion system designed specifically for handling transparent objects, addresses the challenges faced by camera-only approaches in suboptimal conditions. FuseGrasp significantly improves depth completion and grasping success rates by leveraging the robotic arm's motion to capture high-quality  mmWave radar images and employing a deep neural network to fuse RGB-D and radar imagery. To overcome the lack of radar image datasets for transparent objects, we propose a two-stage training approach: pre-training on a large RGB-D dataset and fine-tuning on a small self-built RGB-D-Radar dataset. Additionally, FuseGrasp identifies object materials, enabling the robotic arm to adjust its grip force accordingly. Extensive real-world trials demonstrate that FuseGrasp's effectiveness in detecting and handling transparent objects, with a 45.9\% improvement in grasping success rates compared to camera-guided methods.

\bibliographystyle{IEEEtran}
\bibliography{ref}

\newpage
\appendices
\section{Radar Sensor Selection}
To select an appropriate sensor for imaging transparent objects, we conducted a preliminary experiment comparing two commonly used commodity-grade radar sensors: impulse-radio ultra-wideband (IR-UWB) and mmWave radars. The IR-UWB radar operates at a center frequency of 7.29 GHz with a bandwidth of 1.5 GHz, and the mmWave radar operates at a center frequency of 60 GHz with a bandwidth of 4 GHz \cite{TI, Novelda, Novelda_paper}. In the experiment, two glass cups were placed at the center of a table, and a robotic arm was programmed to swipe over the table with a consistent velocity and trajectory. Each radar device was attached to the robotic arm in the same position, and the received raw signals were recorded.

\begin{figure}[htbp]
	\centering	
	\includegraphics[width = 0.95\linewidth]{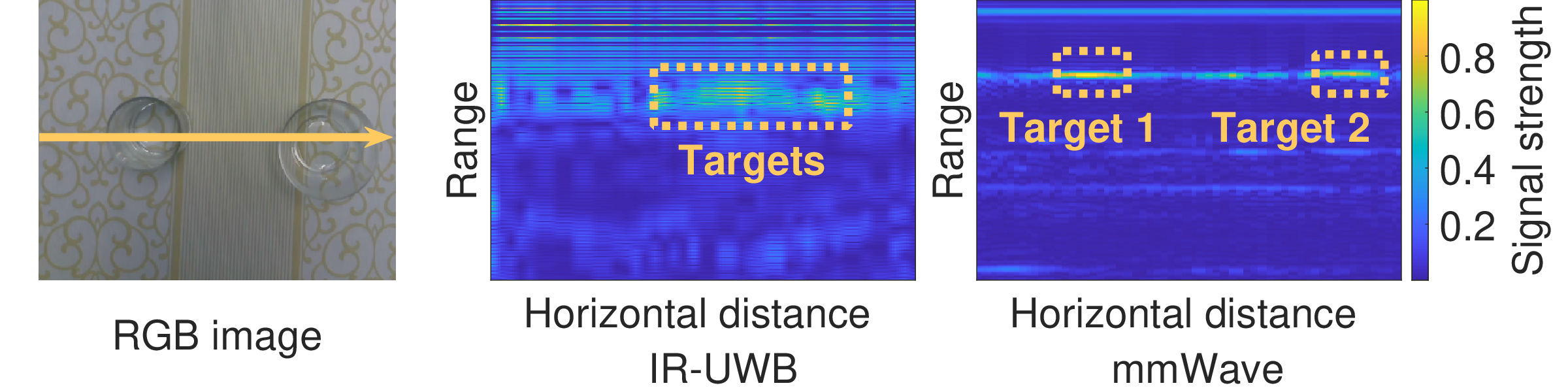} 
	\caption{\textbf{\emph{Left.}} RGB image of two transparent objects placed on the table, in which the arrow is the radar scanning path.
		\textbf{\emph{Middle.}} IR-UWB cannot distinguish two objects clearly. 
		\textbf{\emph{Right.}} mmWave radar can highlight the two regions corresponding to the two objects.}
	\label{Preliminary study case}
\end{figure}

Our observations revealed that while the IR-UWB radar could detect the presence of the targets, excessive diffraction impeded its ability to focus on and distinguish between the two separate objects (Fig. \ref{Preliminary study case}, {middle}). Furthermore, the spatial resolution of the IR-UWB radar was limited by its 1.5 GHz bandwidth, hindering fine-grained imaging. In contrast, thanks to their low penetration capabilities, 60 GHz mmWave signals effectively detected the transparent objects. The signal bandwidth of up to 4 GHz also provided a solid foundation for subsequent radar imaging. Based on these findings, we concluded that mmWave radar outperforms IR-UWB in detecting and imaging transparent objects. However, as the raw received signal only indicates the presence of a target, additional imaging algorithms are required to obtain specific geometric information.

\section{A Summary of the Radar Data Acquisition and Calibration Pipeline}
The radar data acquisition and calibration pipeline in FuseGrasp comprises three main steps: (1) radar data acquisition, (2) radar signal calibration, and (3) workspace background cancellation. In the following, we summarize and provide additional details about the entire pipeline to enhance the readers' understanding.

\begin{figure}[htbp]
	\centering
	\includegraphics[width=0.8\linewidth]{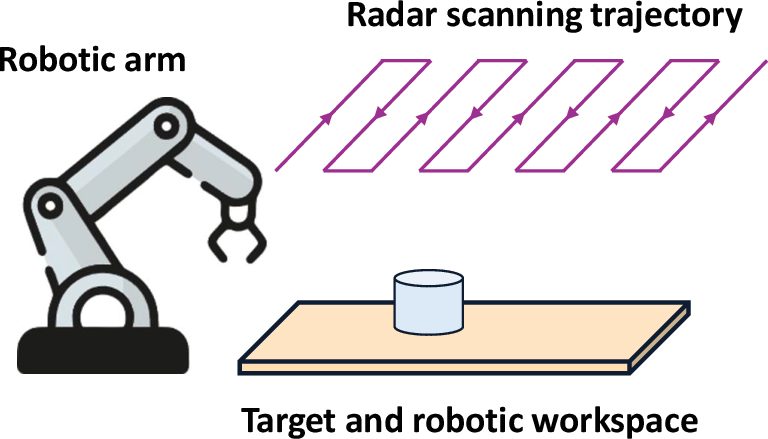}
	\caption{The pre-designed radar scanning trajectory.}
	\label{Scanning_trajectory}
\end{figure}

During the radar data acquisition process, we first activated the radar board to ensure it was functioning correctly and outputting the received signal as expected. Subsequently, the robotic arm automatically sweeps across the workspace table along a pre-designed trajectory, as shown in Fig. \ref{Scanning_trajectory}. This movement generates a uniform antenna planar array with 161 × 84 virtual antenna elements along the horizontal and vertical axes. For each location coordinate of the virtual antenna array, as well as its corresponding timestamp in the robotic arm's position data stream, we find the nearest timestamp in the radar data stream to synchronize the data streams. Once synchronized, the received raw radar data are transmitted to the robotic imaging module for constructing high-quality radar images. Further details on data stream synchronization can be found in Section III-A (1) of the main body of our paper.

\begin{figure}
	\centering	
	\subfigure[The RGB and raw radar image of two transparent objects.]{
		\includegraphics[width = 0.9\linewidth]{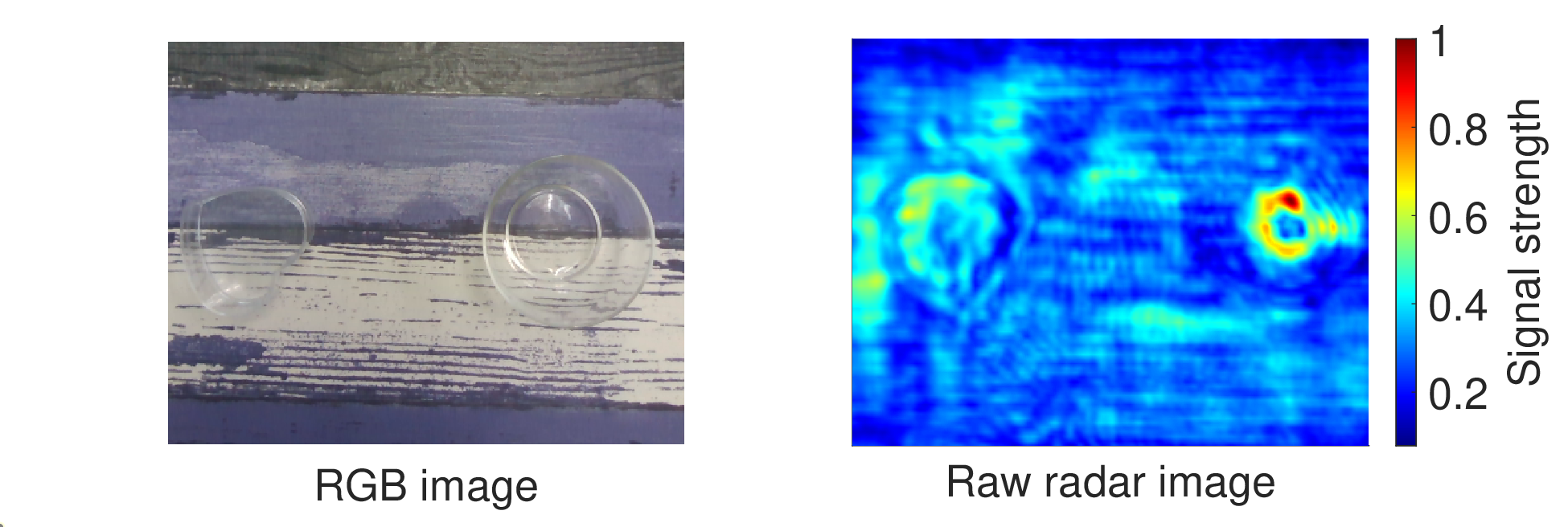}} 
	\label{desk_sar_rawdata}
	
	\vfill
	\subfigure[The enhanced radar image after signal calibration and background cancellation.]{
		\includegraphics[width = 0.9\linewidth]{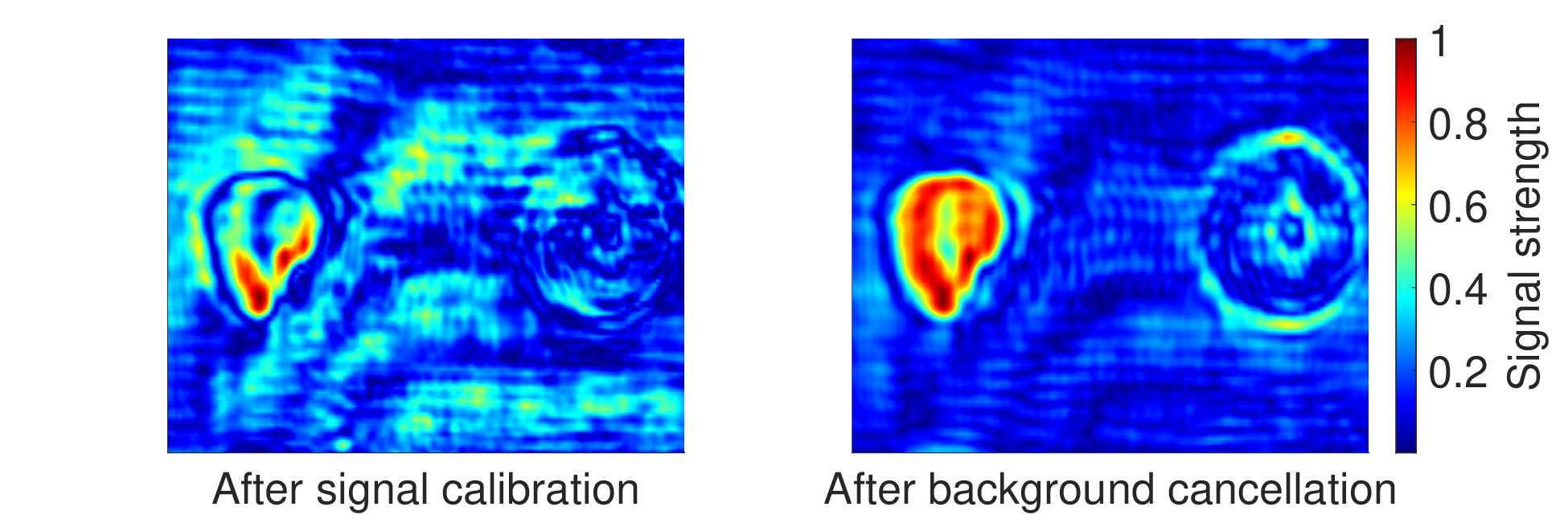}}
	\label{desk_sar}
	
	\caption{Results of two transparent objects placed on the table.}
	\label{desk_sar_pipeline}
\end{figure}

However, in real-world scenarios, factors such as channel mismatches or hardware offsets can introduce phase errors, leading to radar image blurring. Additionally, background noise further degrades the quality of radar imaging. For instance, Fig. \ref{desk_sar_pipeline}a shows an RGB image alongside the raw radar image, where a transparent heart-shaped box and a circular bowl were placed on the robotic workspace. In the original radar results, the outlines of two transparent targets are not only blurry but also overshadowed by signals reflected from the workspace. To address these issues, we designed signal calibration and background cancellation schemes. These methods aim to mitigate phase errors and reduce the impact of background noise, thereby improving the clarity and accuracy of the radar images.

In the radar signal calibration step, our objective is to accurately collect and eliminate the phase offset in the received signals. To achieve this, we used a static point target with high reflectivity as a reference to construct an ideal signal model. Using this model, we estimated the phase bias present in the raw data captured by the mmWave radar and effectively removed it, ensuring accurate phase alignment between the received and ideal signals. The radar images after applying signal calibration are shown in Fig. \ref{desk_sar_pipeline}b (left). Compared with the raw radar image in Fig.~\ref{desk_sar_pipeline}a (right), the geometrical shapes of the two transparent objects have become clearer. Additional mathematical details related to this step can be found in Section III-A (2) of the main body of our paper.

Nevertheless, Fig. \ref{desk_sar_pipeline}b (left) shows that reflected signals from the workspace table are strong and close to the detected targets, making direct cancellation challenging. To address this issue, FuseGrasp employs a simple yet effective method for background cancellation. Specifically, the robotic arm first sweeps the empty workspace to construct a background radar image. This background image is then subtracted from the output radar images. The result of applying background cancellation is shown in Fig. \ref{desk_sar_pipeline}b (right). As demonstrated, both signal calibration and background cancellation are crucial steps in constructing high-quality radar images.

\section{Our RGB-D-Radar Dataset}
\begin{figure}
	\centering	
	\subfigure[Two data samples: normal group (top) and dim group (bottom).]{
		\includegraphics[width = 0.8\linewidth]{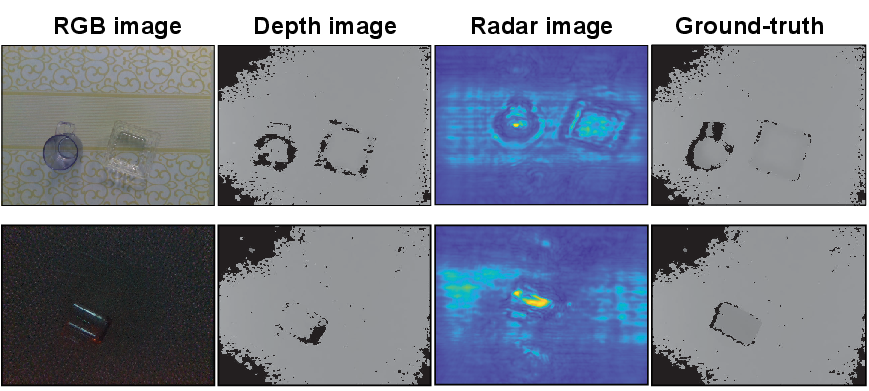}} 
	\label{dataset_samples}
	
	\vfill
	\subfigure[Four different background texture.]{
		\includegraphics[width = 0.8\linewidth]{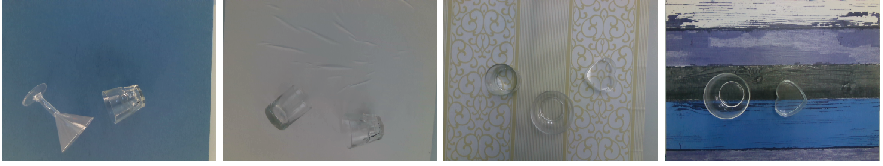}}
	\label{dataset_texture}
	\vspace{-1em}
	
	\caption{Samples and backgrounds of self-built dataset.}
	\label{self_dataset}
\end{figure}

We made a new dataset of transparent objects consisting of RGB, depth, and mmWave radar images. A summary of our dataset is provided in Table \ref{self-built}. Our collection includes a total of 600 realistic RGB-D-Radar data pairings, spread across four distinctive background settings (see Fig. \ref{self_dataset}b). Each setting is carefully chosen to include at least one transparent object, aiming to faithfully represent various indoor environments where robotic arm manipulations are common. 
To acquire precise ground-truth depth images, we meticulously crafted replicas of the transparent objects, painting them black to emulate their opaque counterparts of identical shape~\cite{cleargrasp}.

\begin{table}
	\caption{Summary of self-built transparent dataset.}
	\centering
	\begin{adjustbox}{max width= 0.45 \textwidth}
		\begin{tabularx}{0.6\textwidth} { 
				>{\centering\arraybackslash}X |
				*{3} {>{\centering\arraybackslash}X }
			}
			\toprule[1.3pt]
			Type & \# Scene settings & \# Data pairs & \# Targets  \\ \hline
			Normal group & 4 & 459 & 12\\ 
			Dim group & 4 & 141 & 10\\ 
			\bottomrule[1.3pt]
		\end{tabularx}
	\end{adjustbox}
	\vspace{-1em}
	\label{self-built}
\end{table}

Our dataset encompasses commonly encountered transparent objects from everyday life, placed on a 50 cm × 50 cm operational table. These objects range from glass and plastic cups to dishes and vases. In an effort to understand the effects of lighting on sensor performance, data collection sessions were scheduled at different times of the day, capturing the nuances of varying ambient lighting conditions. The dataset is bifurcated into two subsets to evaluate sensor capabilities under diverse lighting scenarios: the `normal' group, which includes images taken under standard indoor and sunlight conditions, and the `dim' group, which contains images captured under significantly reduced lighting conditions to challenge the limits of optical sensing technologies. We note that the 2-D radar images were exclusively captured at the same elevation as the table surface to ensure consistency. We have included a selection of samples from our dataset in Fig. \ref{self_dataset}a, showcasing the dataset's potential utility for enhancing robotic perception of transparent objects.

\end{document}